National Research Council Canada
Conseil national de recherches Canada

Institute for Information Technology
Institut de technologie de l'information

# *Learning to Extract Keyphrases from Text*

P. Turney

## Abstract

Many academic journals ask their authors to provide a list of about five to fifteen key words, to appear on the first page of each article. Since these key words are often phrases of two or more words, we prefer to call them keyphrases. There is a surprisingly wide variety of tasks for which keyphrases are useful, as we discuss in this paper. Recent commercial software, such as Microsoft's Word 97 and Verity's Search 97, includes algorithms that automatically extract keyphrases from documents.[1] In this paper, we approach the problem of automatically extracting keyphrases from text as a supervised learning task. We treat a document as a set of phrases, which the learning algorithm must learn to classify as positive or negative examples of keyphrases. Our first set of experiments applies the C4.5 decision tree induction algorithm to this learning task. The second set of experiments applies the GenEx algorithm to the task. We developed the GenEx algorithm specifically for this task. The third set of experiments examines the performance of GenEx on the task of metadata generation, relative to the performance of Microsoft's Word 97. The fourth and final set of experiments investigates the performance of GenEx on the task of highlighting, relative to Verity's Search 97. The experimental results support the claim that a specialized learning algorithm (GenEx) can generate better keyphrases than a general-purpose learning algorithm (C4.5) and the non-learning algorithms that are used in commercial software (Word 97 and Search 97).

## Contents

# 1. *Introduction*

Many journals ask their authors to provide a list of *key words* for their articles. We call these *keyphrases*, rather than key words, because they are often phrases of two or more words, rather than single words. We define a *keyphrase list* as a short list of phrases (typically five to fifteen noun phrases) that capture the main topics discussed in a given document. This paper is concerned with the automatic extraction of keyphrases from text.

Keyphrases are meant to serve multiple goals. For example, (1) when they are printed on the first page of a journal article, the goal is summarization. In this case, the keyphrases are like an extreme form of abstract. They enable the reader to quickly determine whether the given article is in the reader's fields of interest. (2) When they are printed in the cumulative index for a journal, the goal is indexing. They enable the reader to quickly find a relevant article when the reader has a specific need. (3) When a search engine form has a field labelled "key words", the goal is to enable the reader to make the search more precise. A search for documents that match a given query term in the "key word" field will yield a smaller, higher quality list of hits than a search for the same term in the full text of the documents. Keyphrases can serve these diverse goals and others, because the goals share the common requirement for a short list of phrases that capture the main topics discussed in the documents.

In Section 2, we present five different software applications for keyphrases. (1) Keyphrases are a valuable part of document metadata (meta-information for document management). (2) Documents can be skimmed more easily when keyphrases are highlighted. (3) Keyphrases can be used as index terms for searching in document collections. (4) Keyphrases can be used as suggestions for refining a query to a search engine. (5) Keyphrases can be used to analyze usage patterns in web server logs. These examples motivate our research by showing the utility of keyphrases.

There is a need for tools that can automatically create keyphrases. Although keyphrases are very useful, only a small minority of the many documents that are available on-line today have keyphrases. In HTML, the META tag enables authors to include keywords in their documents, but the common practice today is to use the keywords in the META tag to bias search engines. Authors use the keyword META tag to increase the likelihood that their web pages will appear, with a high rank, when people enter queries in web search engines. Therefore the keyword tag is typically filled with a huge list of barely relevant terms. These terms are not intended for human consumption. The demand for good, human-readable keyphrases far exceeds the current supply.

We define *automatic keyphrase extraction* as the automatic selection of important, topical phrases from within the body of a document. Automatic keyphrase extraction is a special case of the more general task of *automatic keyphrase generation*, in which the generated phrases do not necessarily appear in the body of the given document. Section 3 discusses criteria for measuring the performance of automatic keyphrase extraction algorithms. In the experiments in this paper, we measure the performance by comparing machine-generated keyphrases with human-generated keyphrases. In our document collections, 65% to 90% of the author's keyphrases appear somewhere in the body of the document. An ideal keyphrase extraction algorithm could (in principle) generate phrases that match up to 90% of the author's keyphrases.

We discuss related work by other researchers in Section 4. Perhaps the most closely related work involves the problem of *automatic index generation* (Fagan, 1987; Salton, 1988; Ginsberg, 1993; Nakagawa, 1997; Leung and Kan, 1997). Although keyphrases may be used in an index, keyphrases have other applications, beyond indexing. The main difference between a keyphrase list and an index is length. Because a keyphrase list is relatively short, it must contain only the most important, topical phrases for a given document. Because an index is relatively long, it can contain many less important, less topical phrases. Also, a keyphrase list can be read and judged in seconds, but an index might never be read in its entirety. Automatic keyphrase extraction is, in many ways, a more demanding task than automatic index generation.

Keyphrase extraction is also distinct from *information extraction*, the task that has been studied in depth in the *Message Understanding Conferences* (MUC-3, 1991; MUC-4, 1992; MUC-5, 1993; MUC-6, 1995). Information extraction involves extracting specific types of task-dependent information. For example, given a collection of news reports on terrorist attacks, information extraction involves finding specific kinds of information, such as the name of the terrorist organization, the names of the victims, and the type of incident (e.g., kidnapping, murder, bombing). In contrast, keyphrase extraction is not specific. The goal in keyphrase extraction is to produce topical phrases, for any type of factual document (automatic keyphrase extraction is not likely to work well with poetry).

The experiments in this paper use five collections of documents, with a combined total of 652 documents. Each document has an associated list of human-generated keyphrases. We use 290 documents for training the learning algorithms and 362 for testing. The collections are presented in detail in Section 5.

We approach automatic keyphrase extraction as a supervised learning task. We treat a document as a set of phrases, which must be classified as either positive or negative examples of keyphrases. This is the classical machine learning problem of *learning from examples.* In Section 6, we describe how we apply the C4.5 decision tree induction algorithm to this task (Quinlan, 1993). There are several unusual aspects to this classification problem. For example, the positive examples constitute only 0.2% to 2.4% of the total number of examples. C4.5 is typically applied to more balanced class distributions.

In our first set of experiments (Section 7), we evaluate different ways to apply C4.5. In preliminary experiments with the training documents, we found that *bagging* seemed to improve the performance of C4.5 (Breiman, 1996a, 1996b; Quinlan, 1996). Bagging works by generating many different decision trees and allowing them to vote on the classification of each example. We experimented with different numbers of trees and different techniques for sampling the training data. The experiments support the hypothesis that bagging improves the performance of C4.5 when applied to automatic keyphrase extraction.

During the course of our experiments with C4.5, we came to believe that a specialized algorithm, developed specifically for learning to extract keyphrases from text, might achieve better results than a general-purpose learning algorithm, such as C4.5. Section 8 introduces the *GenEx* algorithm. *GenEx* is a hybrid of the *Genitor* steady-state genetic algorithm (Whitley, 1989) and the *Extractor* parameterized keyphrase extraction algorithm (NRC, patent pending). Extractor works by assigning a numerical score to the phrases in the input document. The final output of Extractor is essentially a list of the highest scoring phrases. The behaviour of the scoring function is determined by a dozen numerical parameters. Genitor tunes the setting of these parameters, to maximize the performance of Extractor on a given set of training examples.

The second set of experiments (Section 9) supports the hypothesis that a specialized learning algorithm (GenEx) can generate better keyphrases than a general-purpose learning algorithm (C4.5). This is not surprising, because a strongly biased, specialized learning algorithm should be able to perform better than a weakly biased, general-purpose learning algorithm, when the bias is suitable for the given learning problem. However, the results are interesting, because they support the hypothesis that the bias that is built into GenEx is suitable for its intended task.

Although the experiments show that GenEx extracts better keyphrases than C4.5, it is not clear from the experiments whether either algorithm works well enough to be used in commercial software. Therefore, in Section 10, we use GenEx to generate keyphrase metadata and we compare the resulting metadata with the metadata generated by the AutoSummarize feature in Microsoft's Word 97. We find that GenEx metadata is significantly closer to human-generated metadata. The experiments support the claim that the performance of GenEx is sufficiently good for commercial applications in automatic metadata generation.

In Section 11, we apply GenEx to the task of highlighting text to facilitate skimming and we compare the highlighted terms with the terms selected by Verity's Search 97. We find that GenEx selects terms that are more similar to human-generated keyphrases than the terms selected by Verity's Search 97. The experiments support the claim that the performance of GenEx is sufficiently good for commercial applications in automatic highlighting.

Section 12 presents the current status of our work with GenEx and our plans for future work. We conclude (in Section 13) that it is feasible to learn to extract keyphrases from text. It appears that a specialized learning algorithm can achieve better performance than current non-learning commercial algorithms for automatically extracting keyphrases from text.

## 2. *Applications for Keyphrases*

There are many potential applications for keyphrases. In this section, we discuss five actual applications. The variety of applications is an important motivation for our research in learning to automatically extract keyphrases from documents.

### 2.1 Keyphrases for Metadata

With the growth of the Internet and corporate intranets, the problems of document management have also grown. Many researchers believe that metadata is essential to address the problems of document management. Metadata is meta-information about a document or set of documents. There are several standards for document metadata, including the Dublin Core Metadata Element Set (championed by the US Online Computer Library Center), the MARC (Machine-Readable Cataloging) format (maintained by the US Library of Congress), the GILS (Government Information Locator Service) standard (from the US Office of Social and Economic Data Analysis), and the CSDGM (Content Standards for Digital Geospatial Metadata) standard (from the US Federal Geographic Data Committee). All of these standards include a field for keyphrases (although they have different names for this field).

Some software tools for editing documents now include support for metadata creation. For example, Microsoft's Word 97 can store metadata for a document and it can automatically generate keyphrase metadata. Figure 1 shows an example of the "File Properties" metadata template in Microsoft's Word 97. In this illustration, the Keywords and Comments fields were filled automatically, as a side-effect of selecting AutoSummarize from the Tools menu. The Title and Author fields were filled by hand. A good algorithm for automatically generating keyphrases can be very helpful for metadata creation.

### 2.2 Keyphrases for Highlighting

When we skim a document, we scan for keyphrases, to quickly determine the topic of the document. *Highlighting* is the practice of emphasizing keyphrases and key passages (e.g., sentences or paragraphs) by underlining the key text, using a special font, or marking the key text with a special colour. The purpose of highlighting is to facilitate skimming.

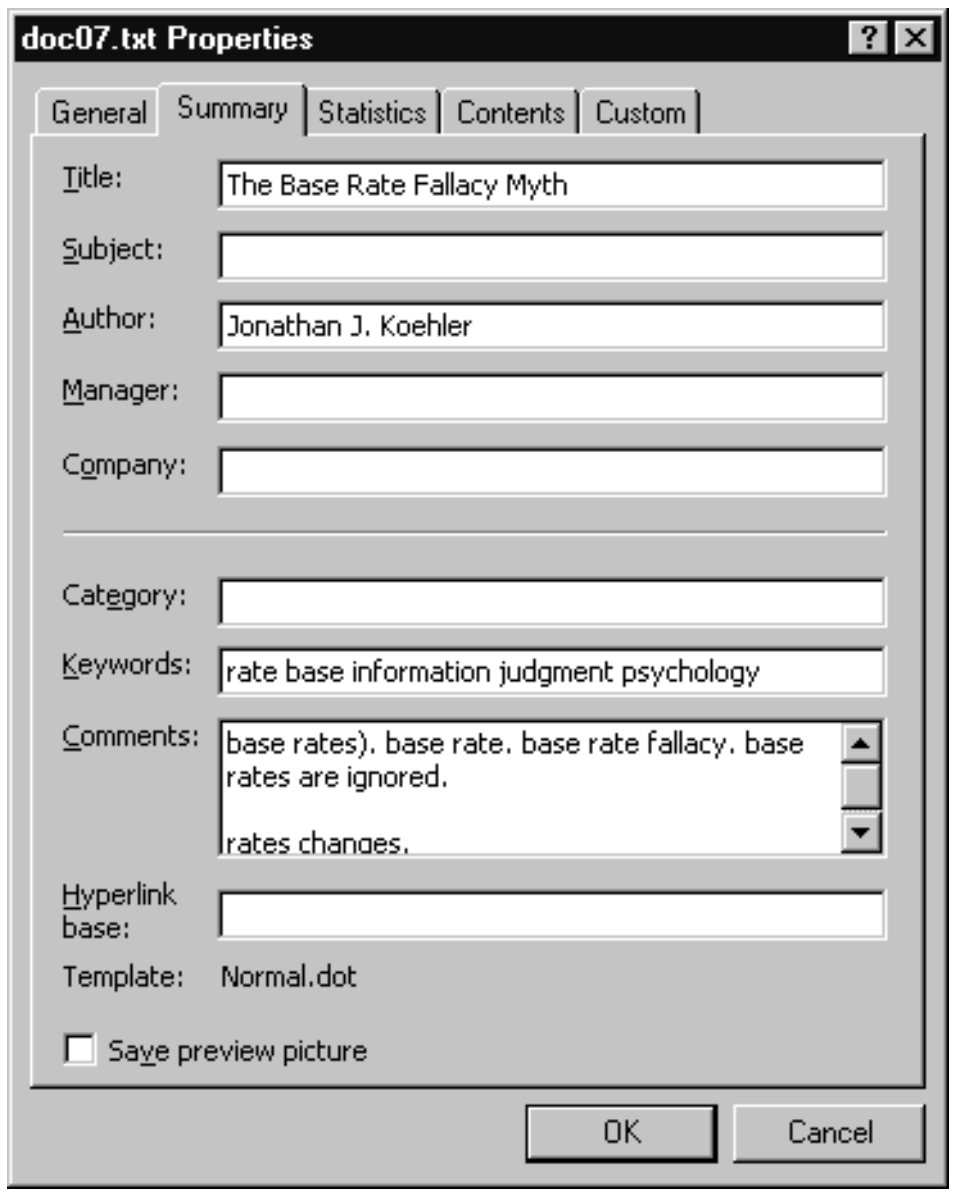

Figure 1: **Keyphrases for metadata:** This pop-up window shows an example of the "File Properties" metadata in Microsoft's Word 97.

Some software tools for document management now support skimming by automatically highlighting keyphrases. For example, Verity's Search 97 is a commercial search engine. When the user enters a query in Verity's Search 97, the list of matching documents (the "hit list") can contain automatically generated summaries of each document, using Search 97's Summarize feature. Figure 2 shows an example of a hit list in Search 97. Keyphrases within each summary are automatically selected and highlighted with a bold font. A good algorithm for automatically generating keyphrases can be very helpful for automatic highlighting.

## 2.3 Keyphrases for Indexing

An alphabetical list of keyphrases, taken from a collection of documents or from parts of a single long document (e.g., chapters in a book), can serve as an index. Figure 3 shows a Java applet (developed at NRC by Joel Martin) for manipulating an alphabetical index. In this example, the document collection is around 100 articles from the *Journal of Artificial Intelligence Research.* The alphabetical list of keyphrases was generated using GenEx. In the first window ("1. Type here to narrow phrase list."), the user has entered part of a word, "learni". In the second window, the Java applet lists all of the machine-generated keyphrases in which one of the words in the phrase begins with "learni". The user has selected the phrase "multi-agent reinforcement learning" from this list. The third window shows the title of the article from which this keyphrase was extracted. If the user selects the title, the Java applet will display the abstract of the paper.

## 2.4 Keyphrases for Interactive Query Refinement

Using a search engine is often an iterative process. The user enters a query, examines the resulting hit list, modifies the query, then tries again. Most search engines do not have any special features that support the iterative aspect of searching. Figure 4 shows a search engine (developed at

## 2. *Applications for Keyphrases*

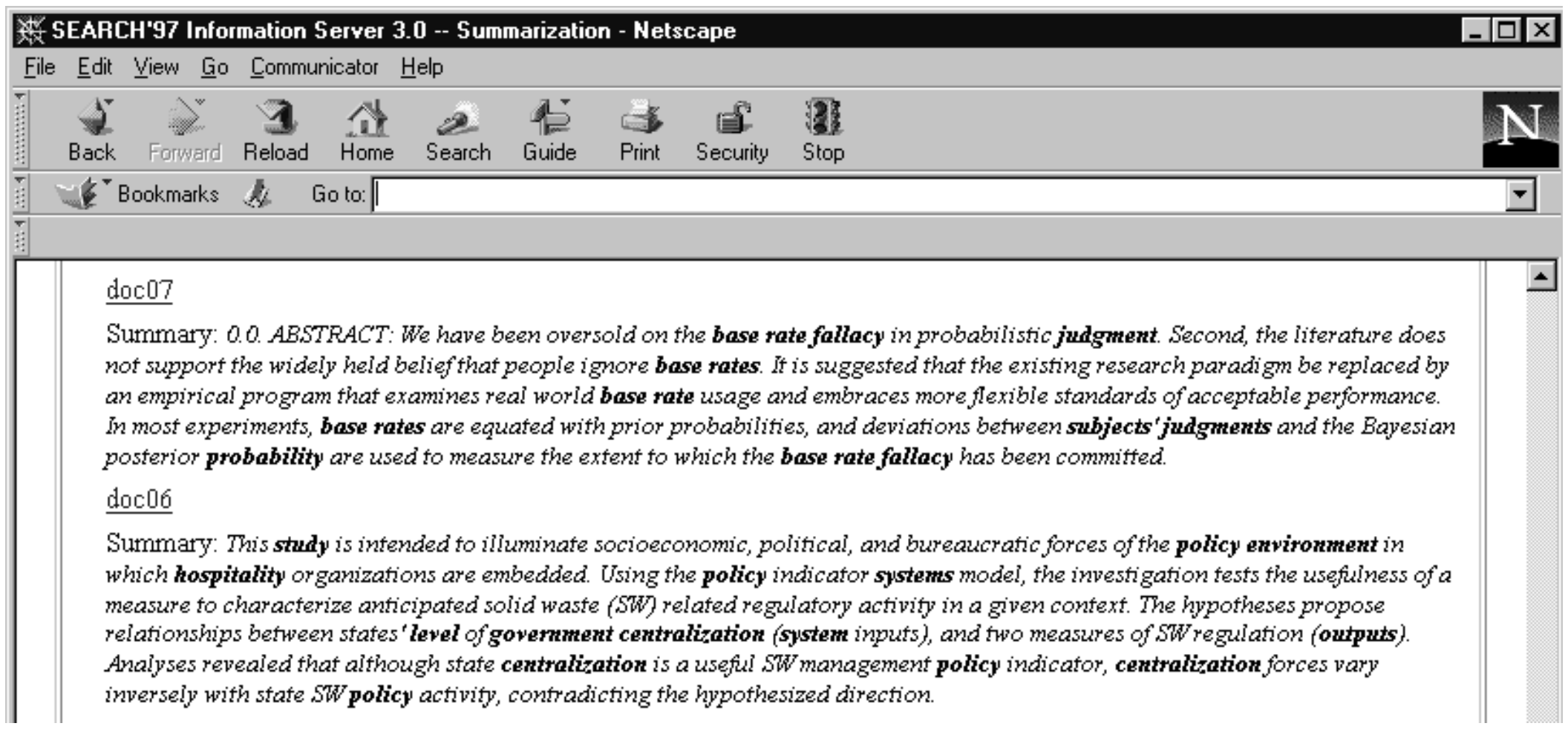

Figure 2: **Keyphrases for highlighting:** In Verity's Search 97, keyphrases within each summary are automatically selected and highlighted with a bold font.

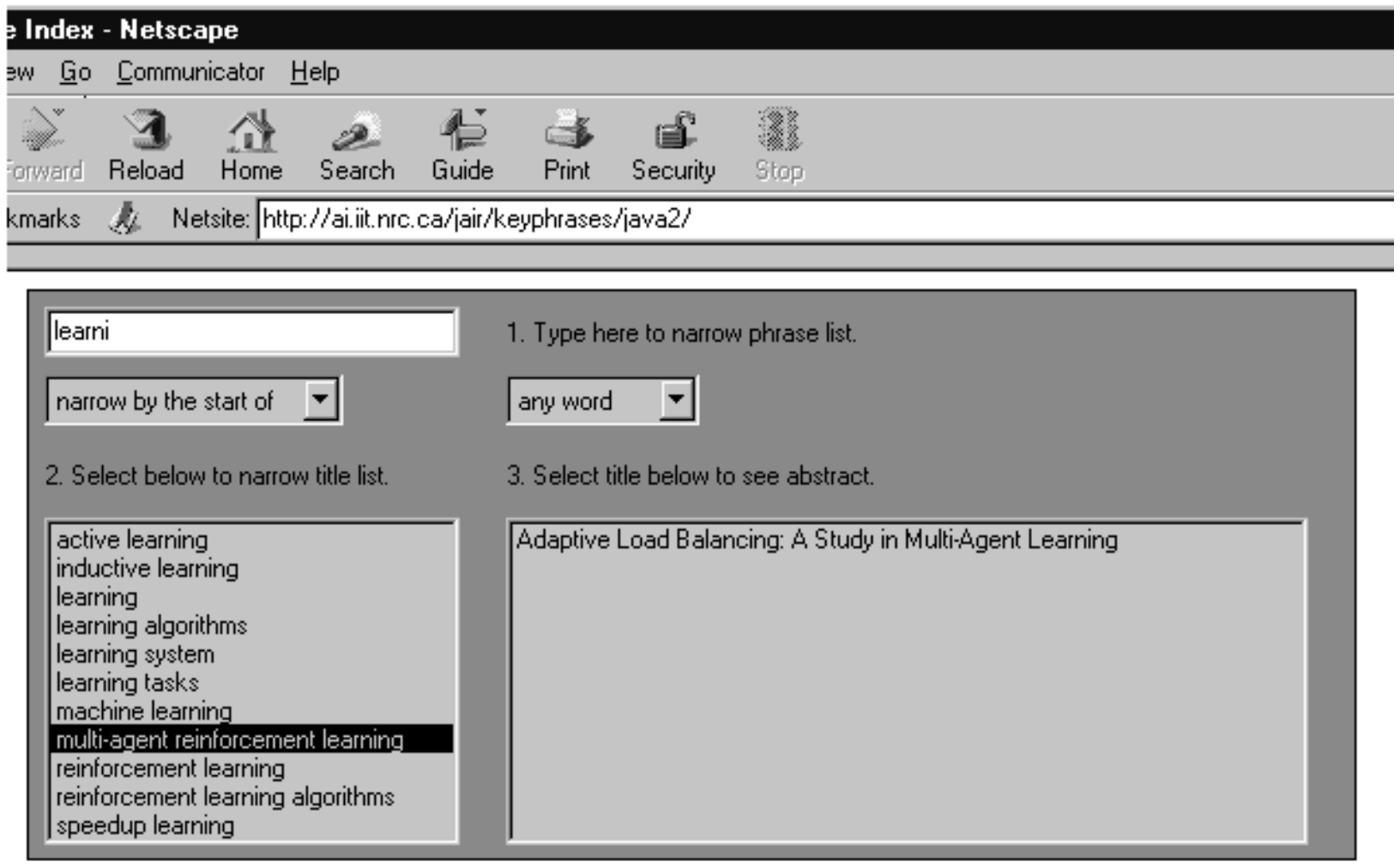

Figure 3: **Keyphrases for indexing:** This Java applet lets users search for journal articles by manipulating a list of automatically generated keyphrases.

NRC) for searching for journal articles in the *Journal of Artificial Intelligence Research*. This search engine was developed to support interactive query refinement. In Figure 4, the user has entered the query "learning" in the top frame. The left frame shows the matching documents and the right frame lists suggestions for narrowing the original query. These suggestions are keyphrases extracted by GenEx from the documents that are listed in the left frame. The query terms are combined by conjunction, so the hit list (in the left frame) becomes smaller with each iteration. However, adding the suggested terms (in the right frame) will never result in an empty hit list, because the terms necessarily appear in at least one of the documents in the hit list.

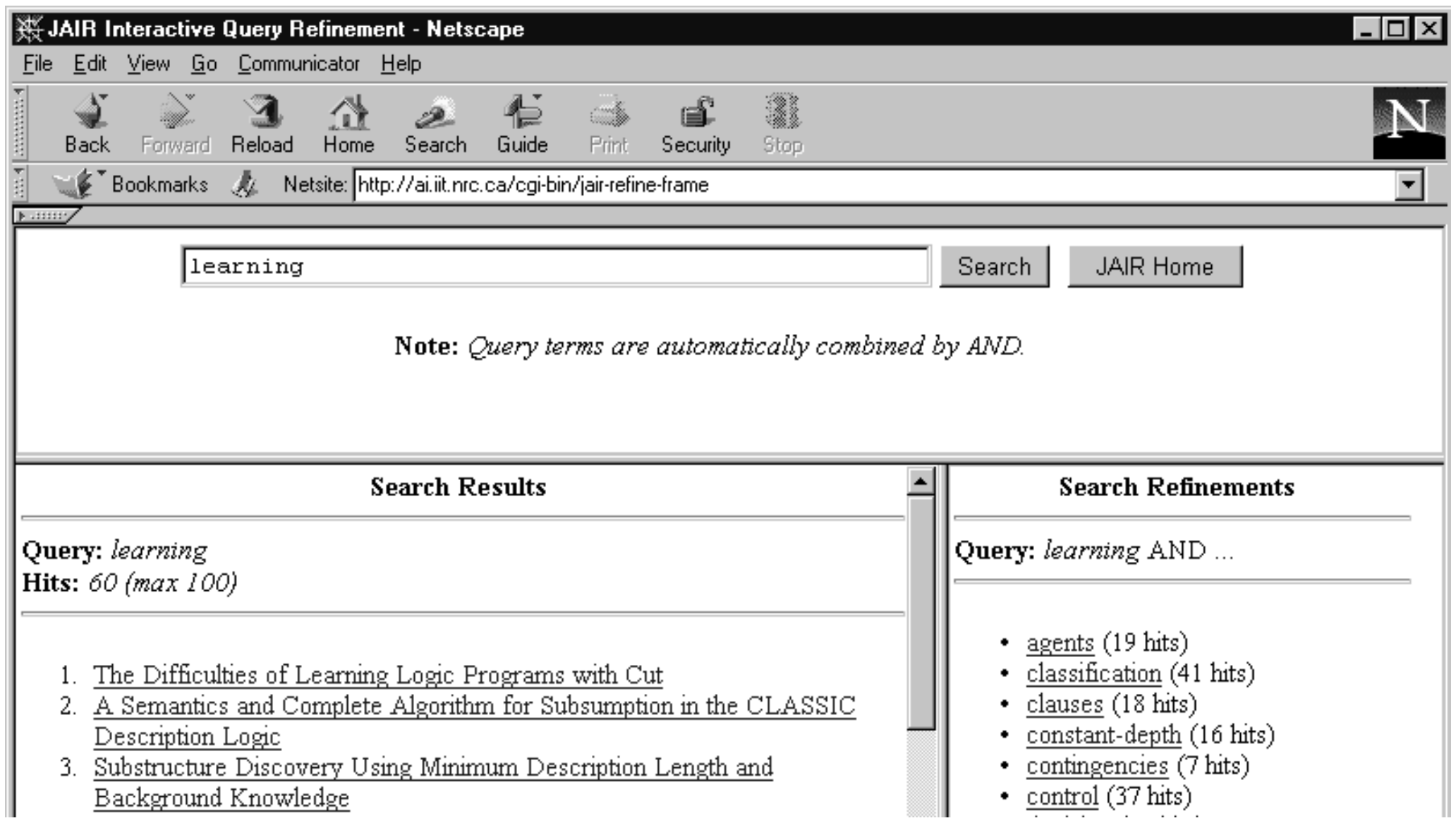

Figure 4: **Keyphrases for interactive query refinement:** The suggested query refinements in the right frame are based on keyphrases that are automatically extracted from the documents listed in the left frame.

### 2.5 Keyphrases for Web Log Analysis

Web site managers often want to know what visitors to their site are seeking. Most web servers have log files that record information about visitors, including the Internet address of the client machine, the file that was requested by the client, and the date and time of the request. There are several commercial products that analyze these logs for web site managers. Typically these tools will give a summary of general traffic patterns and produce an ordered list of the most popular files on the web site.

Figure 5 shows the output of a web log analysis program that uses keyphrases (developed at NRC). Instead of producing an ordered list of the most popular files on the web site, this tool produces a list of the most popular keyphrases on the site. This gives web site managers insight into which *topics* on the web site are most popular.

## 3. *Measuring the Performance of Keyphrase Extraction Algorithms*

We measure the performance of keyphrase extraction algorithms by comparing their output to handmade keyphrases. The performance measure is based on the number of matches between the machine-generated phrases and the human-generated phrases. In the following subsections, we define what we mean by *matching* phrases and we describe how the performance measure is calculated from the number of matches.

### 3.1 Criteria for Matching Phrases

If an author suggests the keyphrase “neural network” and a keyphrase generation algorithm suggests the keyphrase “neural networks”, we want to count this as a match, although one phrase is singular and the other is plural. On the other hand, if the author suggests "neural networks" and the algorithm suggests "networks", we do not want to count this as a match, since there are many different kinds of networks.

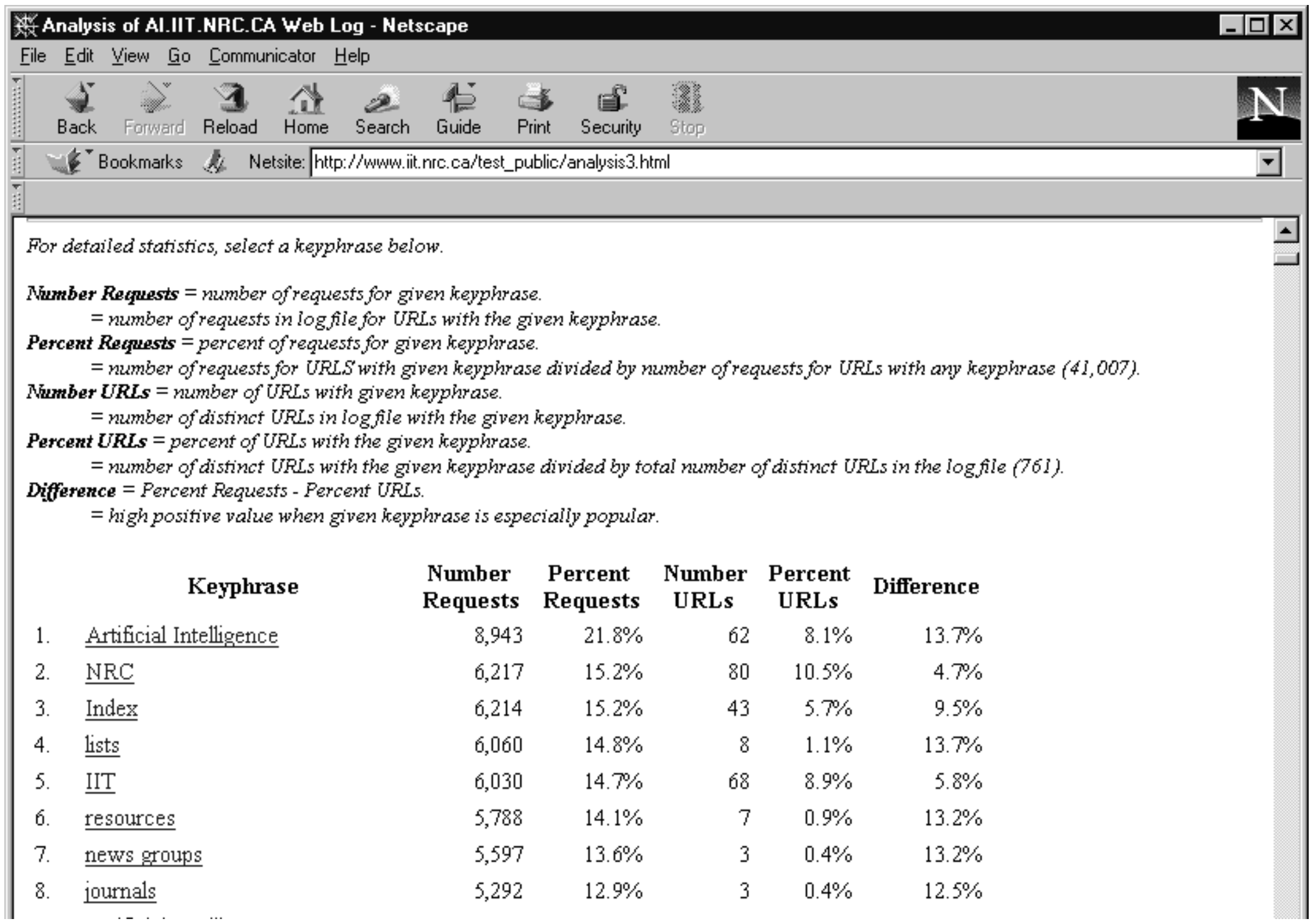

Figure 5: **Keyphrases for web log analysis:** Traffic records in a web server log are analyzed here by looking at the frequency of keyphrases, instead of the frequency of URLs. The keyphrases are automatically extracted from the web pages on the server.

In the experiments that follow, we say that a handmade keyphrase matches a machine-generated keyphrase when they correspond to the same sequence of stems. A stem is what remains when we remove the suffixes from a word. By this definition, "neural networks" matches "neural network", but it does not match "networks". The order in the sequence is important, so "helicopter skiing" does not match "skiing helicopter". To be more precise about our criteria for matching phrases, we need to say more about how a word is converted to its stem.

## 3.2 Stemming

The Porter (1980) and Lovins (1968) stemming algorithms are the two most popular algorithms for stemming English words.[2] Both algorithms use heuristic rules to remove or transform English suffixes. Another approach to stemming is to use a dictionary that explicitly lists the stem for every word that might be encountered in the given text. Heuristics are usually preferred to a dictionary, due to the labour involved in constructing the dictionary and the computer resources (storage space and execution time) required to use the dictionary.

The Lovins stemmer is more aggressive than the Porter stemmer. That is, the Lovins stemmer is more likely to map two words to the same stem, but it is also more likely to make mistakes (Krovetz, 1993). For example, the Lovins stemmer correctly maps "psychology" and "psychologist" to the same stem, "psycholog", but the Porter stemmer incorrectly maps them to different stems, "psychologi" and "psychologist". On the other hand, the Porter stemmer correctly maps "police" and "policy" to different stems, "polic" and "polici", but the Lovins stemmer incorrectly maps them to the same stem, "polic".

We have found that aggressive stemming is better for keyphrase extraction than conservative stemming. In our experiments, we have used an aggressive stemming algorithm that we call the Iterated Lovins stemmer. The algorithm repeatedly applies the Lovins stemmer, until the word stops changing. For example, given "incredible" as input, the Lovins stemmer generates "incred" as output. Given "incred" as input, it generates "incr" as output. With "incr" as input, the output is also "incr". Thus the Iterated Lovins algorithm, given "incredible" as input, generates "incr" as output. Iterating in this manner will necessarily increase (or leave unchanged) the aggressiveness of any stemmer.

## 3.3 Precision and Recall

We may view keyphrase extraction as a classification problem. If we think of a document as a set of words and phrases, then the task is to classify each word or phrase into one of two categories: either it is a keyphrase or it is not a keyphrase. We can evaluate an automatic keyphrase extraction algorithm by the degree to which its classifications correspond to the human-generated classifications. The outcome of applying a keyphrase extraction algorithm to a corpus can be neatly summarized with a *confusion matrix*, as in Table 1. The variable $a$ represents the number of times that the human-generated keyphrase matches a machine-generated keyphrase. The variable $d$ represents the number of times that the human and the machine agree that a phrase is not a keyphrase. The variables $b$ and $c$ represent the number of times that the human and the machine disagree on the classification of a phrase.

Table 1: The confusion matrix for keyphrase classification.

| | Classified as a Keyphrase by the Human | Classified as Not a Keyphrase by the Human |
|---|---|---|
| Classified as a Keyphrase by the Machine | $a$ | $b$ |
| Classified as Not a Keyphrase by the Machine | $c$ | $d$ |

We consider both "neural network" and "neural networks" to be matches for the phrase "neural network". Therefore it is better to think of the task as classification of *stemmed phrases*, rather than classification of *whole phrases*. The Iterated Lovins stemmer transforms both whole phrases "neural network" and "neural networks" into the stemmed phrase "neur network". If a person suggests that "neural networks" is a good keyphrase for an article, we interpret that as classifying the stemmed phrase "neur network" as a keyphrase.

We would like to have a single number that represents the performance of a keyphrase extraction algorithm. In other words, we would like a suitable function that maps $a$, $b$, $c$, and $d$ to a single value. It is common to use *accuracy* to reduce a confusion matrix to a single value:

$$\text{accuracy} = \frac{a + d}{a + b + c + d} \quad (1)$$

Unfortunately, there are some problems with using accuracy here. One problem is that, because the class distribution is highly skewed (there are many more negative examples — phrases that are not keyphrases — than positive examples), we can achieve very high accuracy by always guessing the majority class. That is, if a trivial keyphrase extraction algorithm always generates an empty set of keyphrases ($a = b = 0$), for any input document, its accuracy would typically be above 99%.

Researchers in information retrieval use *precision* and *recall* to evaluate the performance of a search engine:

$$\text{precision} = \frac{a}{a+b} \tag{2}$$

$$\text{recall} = \frac{a}{a+c} \tag{3}$$

Precision is an estimate of the probability that, if a given search engine classifies a document as relevant to a user’s query, then it really is relevant. Recall is an estimate of the probability that, if a document is relevant to a user’s query, then a given search engine will classify it as relevant.

There is a well-known trade-off between precision and recall. We can optimize one at the expense of the other. For example, if we guess that the entire document collection is relevant, then our recall is necessarily 100%, but our precision will be close to 0%. On the other hand, if we take the document that we are most confident is relevant and guess that only this single document is relevant to the user’s query, then our precision might be 100%, but our recall would be close to 0%. We want a performance measure that yields a high score only when precision and recall are balanced.

In the experiments that follow, we use two methods for measuring performance. Both of these methods are widely used in research in information retrieval. (1) We examine precision with a variety of cut-offs for the number of machine-generated keyphrases ($a + b$). This is our preferred method, because it shows how the performance varies as the user adjusts the desired number of keyphrases. (2) We use the *F-measure* (van Rijsbergen, 1979; Lewis, 1995):

$$\text{F-measure} = \frac{2 \cdot \text{precision} \cdot \text{recall}}{\text{precision} + \text{recall}} = \frac{2a}{2a+b+c} \tag{4}$$

When precision and recall are nearly equal, the F-measure is nearly the same as the average of precision and recall. When precision and recall are not balanced, the F-measure is less than the average.

## 4. *Related Work*

In this section, we discuss some related work. Although there are several papers that discuss automatically extracting important phrases, as far as we know, none of these papers treat this problem as supervised learning from examples.

Krulwich and Burkey (1996) use heuristics to extract significant phrases from a document. The heuristics are based on syntactic clues, such as the use of italics, the presence of phrases in section headers, and the use of acronyms. Their motivation is to produce phrases for use as features when automatically classifying documents. Their algorithm tends to produce a relatively large list of phrases, so it has low precision, and thus low F-measure.

Muñoz (1996) uses an unsupervised learning algorithm to discover two-word keyphrases. The algorithm is based on Adaptive Resonance Theory (ART) neural networks. Muñoz’s algo-rithm tends to produce a large list of phrases, so it has low precision, and thus low F-measure. Also, the algorithm is not applicable to one-word or more-than-two-word keyphrases.

Steier and Belew (1993) use the mutual information statistic to discover two-word keyphrases. This approach has the same limitations as Muñoz (1996), when considered as a keyphrase extraction algorithm: it produces a low precision list of two-word phrases. Steier and Belew (1993) compare the mutual information of word pairs within specific topic areas (e.g., documents concerned with labour relations) and across more general collections (e.g., legal documents). They make the interesting observation that certain phrases that would seem to be highly characteristic of a certain topic area (e.g., “union member” would seem to be characteristic of documents concerned with labour relations) actually have a higher mutual information statistic across more general collections (e.g., “union member” has a higher mutual information across a general legal collection than within the topic area of labour relations).

Several papers explore the task of producing a summary of a document by extracting key sentences from the document (Luhn, 1958; Edmundson, 1969; Marsh *et al.*, 1984; Paice, 1990; Paice and Jones, 1993; Johnson *et al.*, 1993; Salton *et al.*, 1994; Kupiec *et al.*, 1995; Brandow *et al.*, 1995; Jang and Myaeng, 1997). This task is similar to the task of keyphrase extraction, but it is more difficult. The extracted sentences often lack cohesion because anaphoric references are not resolved (Johnson *et al.*, 1993; Brandow *et al.*, 1995). *Anaphors* are pronouns (e.g., “it”, “they”), definite noun phrases (e.g., “the car”), and demonstratives (e.g., “this”, “these”) that refer to previously discussed concepts. When a sentence is extracted out of the context of its neighbouring sentences, it may be impossible or very difficult for the reader of the summary to determine the referents of the anaphors. Johnson *et al.* (1993) attempt to automatically resolve anaphors, but their system tends to produce overly long summaries. Keyphrase extraction avoids this problem because anaphors are not keyphrases.[3] Also, a list of keyphrases has no structure; unlike a list of sentences, a list of keyphrases can be randomly permuted without significant consequences.[4]

Most of these papers on summarization by sentence extraction describe algorithms that are based on manually derived heuristics. The heuristics tend to be effective for the intended domain, but they often do not generalize well to a new domain. Extending the heuristics to a new domain involves a significant amount of manual work. A few of the papers describe learning algorithms, which can be trained by supplying documents with associated target summaries (Kupiec *et al.*, 1995; Jang and Myaeng, 1997). Learning algorithms can be extended to new domains with less work than algorithms that use manually derived heuristics. However, there is still some manual work involved, because the training summaries must be composed of sentences that appear in the document, which means that standard author-supplied abstracts are not suitable. An advantage of keyphrase extraction is that standard author-supplied keyphrases are suitable for training a learning algorithm, because the majority of such keyphrases appear in the bodies of the corresponding documents. Kupiec *et al.* (1995) and Jang and Myaeng (1997) use a Bayesian statistical model to learn how to extract key sentences. A Bayesian approach may be applicable to keyphrase extraction.

Another body of related work addresses the task of *information extraction*. An information extraction system seeks specific information in a document, according to predefined guidelines. The guidelines are specific to a given topic area. For example, if the topic area is news reports of terrorist attacks, the guidelines might specify that the information extraction system should identify (i) the terrorist organization involved in the attack, (ii) the victims of the attack, (iii) the type of attack (kidnapping, murder, etc.), and other information of this type that can be expected in a typical document in the topic area. ARPA has sponsored a series of *Message Understanding*

*Conferences* (MUC-3, 1991; MUC-4, 1992; MUC-5, 1993; MUC-6, 1995), where information extraction systems are evaluated with corpora in various topic areas, including terrorist attacks and corporate mergers.

Most information extraction systems are manually built for a single topic area, which requires a large amount of expert labour. The highest performance at the Fifth Message Understanding Conference (MUC-5, 1993) was achieved at the cost of two years of intense programming effort. However, recent work has demonstrated that a learning algorithm can perform as well as a manually constructed system (Soderland and Lehnert, 1994). Soderland and Lehnert (1994) use decision tree induction as the learning component in their information extraction system. We may view the predefined guidelines for a given topic area as defining a template to be filled in by the information extraction system. In Soderland and Lehnert's (1994) system, each slot in the template is handled by a group of decision trees that have been trained specially for that slot. The nodes in the decision trees are based on syntactical features of the text, such as the presence of certain words.

Information extraction and keyphrase extraction are at opposite ends of a continuum that ranges from detailed, specific, and domain-dependent (information extraction) to condensed, general, and domain-independent (keyphrase extraction). The different ends of this continuum require substantially different algorithms. However, there are intermediate points on this continuum. An example is the task of identifying corporate names in business news. This task was introduced in the Sixth Message Understanding Conference (MUC-6, 1995), where it was called the *Named Entity Recognition* task. The competitors in this task were evaluated using the F-measure. The best system achieved a score of 0.964, which indicates that named entity recognition is easier than keyphrase extraction (Krupka, 1995). This system used hand-crafted linguistic rules to recognize named entities.[5]

Other related work addresses the problem of automatically creating an index (Fagan, 1987; Salton, 1988; Ginsberg, 1993; Nakagawa, 1997; Leung and Kan, 1997). Leung and Kan (1997) provide a good survey of this work. There are two general classes of indexes: indexes that are intended for human readers to browse (often called *back-of-book* indexes) and indexes that are intended for use with information retrieval software (*search engine* indexes). Search engine indexes are not suitable for human browsing, since they usually index every occurrence of every word (excluding stop words, such as “the” and “of”) in the document collection. Back-of-book indexes tend to be much smaller, since they index only important occurrences of interesting words and phrases.

Search engine indexes often contain single words, but not multi-word phrases. Several researchers have experimented with extending search engine indexes with multi-word phrases. The result of these experiments is that multi-word phrases have little impact on the performance of search engines (Fagan, 1987; Croft, 1991). They do not appear to be worth the extra effort required to generate them.

Since we are interested in keyphrases for human browsing, back-of-book indexes are more relevant than search engine indexes. Leung and Kan (1997) address the problem of learning to assign index terms from a controlled vocabulary. This involves building a statistical model for each index term in the controlled vocabulary. The statistical model attempts to capture the syntactic properties that distinguish documents for which the given index term is appropriate from documents for which it is inappropriate. Their results are interesting, but the use of a controlled vocabulary makes it difficult to compare their work with the algorithms we examine here. We studied a small sample of controlled index terms in the INSPEC database, and we found that very few of these terms appear in the bodies of the corresponding documents.[6] It seems that algorithms that are suitable for automatically generating controlled index terms are substantially different from algorithms that are suitable for automatically extracting keyphrases. It is also worth noting that a list of controlled index terms must grow every year, as the body of literature grows, so Leung and Kan's (1997) software would need to be continuously trained.

Nakagawa (1997) automatically extracts simple and compound nouns from technical manuals, to create back-of-book indexes. Each compound noun is scored using a formula that is based on the frequency of its component nouns in the given document. In his experiments, Nakagawa (1997) evaluates his algorithm by comparing human-generated indexes to machine-generated indexes. He uses van Rijsbergen's (1979) E-measure, which is simply 1 minus the F-measure that we use in our experiments. His E-measure, averaged over five different manuals, corresponds to an F-measure of 0.670. This suggests that back-of-book indexes are easier to generate than keyphrases. Two factors that complicate the comparison are that Nakagawa (1997) uses Japanese text, whereas we use English text, and Nakagawa's (1997) human-generated indexes were generated with the assistance of his algorithm, which tends to bias the results in favour of his algorithm.

The main feature that distinguishes a back-of-book index from a keyphrase list is length. As Nakagawa (1997) observes, a document is typically assigned $10^0 \sim 10^1$ keyphrases, but a back-of-book index typically contains $10^2 \sim 10^3$ index terms. Also, keyphrases are usually intended to cover the whole document, but index terms are intended to cover only a small part of a document. A keyphrase extraction algorithm might be used to generate a back-of-book index by breaking a long document into sections of one to three pages each. A back-of-book index generation algorithm might be used to generate keyphrases by selecting index terms that appear on many pages throughout the book. Another distinguishing feature is that a sophisticated back-of-book index is not simply an alphabetical list of terms. There is often a hierarchical structure, where a major index term is followed by an indented list of related minor index terms.

# 5. *The Corpora*

The experiments in this paper are based on five different document collections, listed in Table 2. For each document, there is a target set of keyphrases, generated by hand. In the journal article corpus, the keyphrases were created by the authors of the articles. In the email corpus, a university student created the keyphrases. In the three web page corpora, the keyphrases were created by the authors of the web pages (as far as we know).

## 5.1 The Journal Article Corpus

We selected 75 journal articles from five different journals, listed in Table 3. The full text of each article is available on the web. The authors have supplied keyphrases for each of these articles.

Three of the journals are about cognition (*Psycoloquy, The Neuroscientist, Behavioral & Brain Sciences Preprint Archive*), one is about the hotel industry (*Journal of the International Academy of Hospitality Research*), and one is about chemistry (*Journal of Computer-Aided Molecular Design*). This mix of journals lets us see whether there is significant variation in performance of the keyphrase extraction algorithms among journals in the same field (cognition) and among journals in different fields (cognition, hotel industry, chemistry). Our experience indicates that the chemistry journal is particularly challenging for automatic keyphrase extraction.

Table 4 gives some indication of how the statistics vary among the five journals. Most

## 5. *The Corpora*

Table 2: The five document collections.

| Corpus Name | Description | Number of Training Documents | Number of Testing Documents | Total Number of Documents |
|---|---|---|---|---|
| Journal Articles | articles from five different academic journals | 55 | 20 | 75 |
| Email Messages | email messages from six different NRC employees | 235 | 76 | 311 |
| Aliweb Web Pages | web pages from the Aliweb search engine | 0 | 90 | 90 |
| NASA Web Pages | web pages from NASA's Langley Research Center | 0 | 141 | 141 |
| FIPS Web Pages | web pages from the US Government's Federal Information Processing Standards | 0 | 35 | 35 |
| Total | | 290 | 362 | 652 |

Table 3: Sources for the journal articles.

| Journal Name and URL | Number of Documents |
|---|---|
| *Journal of the International Academy of Hospitality Research* http://borg.lib.vt.edu/ejournals/JIAHR/jiahr.html | 6 |
| *Psycoloquy* http://www.princeton.edu/~harnad/psyc.html | 20 |
| *The Neuroscientist* http://www.theneuroscientist.com/ | 2 |
| *Journal of Computer-Aided Molecular Design* http://www.ibc.wustl.edu/jcamd/ | 14 |
| *Behavioral & Brain Sciences Preprint Archive* http://www.princeton.edu/~harnad/bbs.html | 33 |
| Total | 75 |

authors use one or two words in a keyphrase. Occasionally they will use three words, but only rarely will authors use four or five words. The final column shows, for each journal, the percentage of the authors' keyphrases that appear at least once in the full text of the article (excluding the keyword list, of course).

In the following experiments, *Psycoloquy* was used as the testing set and the remaining journals were used as the training set. We chose this split because it resulted in roughly 75% training cases and 25% testing cases. We did not use a random split, because we wanted to test the ability of the learning algorithms to generalize across journals. A random 75/25 split would most likely have resulted in a training set with samples of articles from all five journals. We wanted the testing set to contain articles from a journal that was not represented in the training set. This resulted in 55 training documents and 20 testing documents.

Table 4: Some statistics for each of the five journals.

| Journal Name | Average Number of ... ± Standard Deviation | | | Percentage of Keyphrases in Full Text |
|---|---|---|---|---|
| | Keyphrases per Document | Words per Keyphrase | Words per Document | |
| *Journal of the International Academy of Hospitality Research* | 6.2 ± 2.6 | 2.0 ± 0.8 | 6,299 ± 2,066 | 70.3% |
| *Psycoloquy* | 8.4 ± 3.1 | 1.5 ± 0.6 | 4,350 ± 2,726 | 74.9% |
| *The Neuroscientist* | 6.0 ± 1.4 | 1.8 ± 1.1 | 7,476 ± 856 | 91.7% |
| *Journal of Computer-Aided Molecular Design* | 4.7 ± 1.4 | 1.9 ± 0.6 | 6,474 ± 2,633 | 78.8% |
| *Behavioral & Brain Sciences Preprint Archive* | 8.4 ± 2.2 | 1.6 ± 0.7 | 17,522 ± 6,911 | 87.4% |
| All Five Journals | 7.5 ± 2.8 | 1.6 ± 0.7 | 10,781 ± 7,807 | 81.6% |

## 5.2 The Email Message Corpus

We collected 311 email messages from six different NRC employees. Most of the messages were incoming mail, both internal NRC mail and external mail. We believe that these messages are representative of typical messages that are exchanged in corporate and institutional environments.

A university student created the keyphrases for the 311 email messages. The instructions to the student were, “Please create a list of key words for each of these email messages by selecting phrases from the body or subject field of each message.” We did not give a definition of “key word”. Instead, we gave the student some examples of journal articles with keyphrases. To avoid biasing the student, we did not explain our algorithms or our experimental designs until after the student had finished creating the keyphrases. Since one individual created all of the keyphrases, this corpus is likely to be more homogenous than the journal article corpus or the web page corpora.

Table 5 shows that there is relatively little variation in the statistical properties of the messages among the six employees. As in the journal article corpus, most keyphrases have one to three words. The keyphrases in the email corpus tend to be slightly longer than the keyphrases in the journal corpus. It seems likely that this is a reflection of the tastes of the student, rather than an intrinsic property of email messages.

In the following experiments, the data were randomly split into testing and training sets, by randomly selecting, for each employee, 75% of the messages for training and 25% for testing. This resulted in 235 training messages and 76 testing messages.

## 5.3 The Aliweb Web Page Corpus

We collected 90 web pages using the Aliweb search engine, a public search engine provided by NEXOR Ltd. in the UK, at http://www.nexor.com/public/aliweb/search/doc/form.html. Most web search engines use a *spider* to collect web pages for their index. A spider is a program that gathers web pages by expanding an initial list of URLs by following the hypertext links on the corresponding web pages. Aliweb is unusual in that it does not use a spider to collect web pages; instead, it has an electronic fill-in form, where people are asked to enter any URLs that they would like to add to the Aliweb index. Among other things, this fill-in form has a field for keyphrases. The keyphrases are stored in the Aliweb index, along with the URLs.

Table 5: Some statistics for each of the six employees.

| Employee | Number of Messages | Average Number of … ± Standard Deviation | | | Percentage of Keyphrases in Full Text |
|---|---|---|---|---|---|
| | | Keyphrases per Document | Words per Keyphrase | Words per Document | |
| #1 | 47 | 6.6 ± 5.0 | 2.0 ± 1.1 | 542 ± 606 | 97.4% |
| #2 | 41 | 7.7 ± 6.7 | 1.7 ± 0.8 | 328 ± 374 | 97.5% |
| #3 | 42 | 4.3 ± 3.8 | 1.5 ± 0.8 | 454 ± 698 | 98.9% |
| #4 | 96 | 3.6 ± 2.1 | 1.8 ± 1.0 | 230 ± 243 | 97.1% |
| #5 | 41 | 4.6 ± 4.3 | 1.6 ± 0.8 | 453 ± 805 | 98.9% |
| #6 | 44 | 3.9 ± 2.5 | 2.1 ± 1.1 | 413 ± 674 | 98.8% |
| All Six Employees | 311 | 4.9 ± 4.3 | 1.8 ± 1.0 | 376 ± 561 | 97.9% |

We did a *substring* search with Aliweb, using the query string “e” (the most common letter in English text) and setting the maximum number of matches at 1000. The intent of this search was to gather a relatively large, random sample of web pages. We looked at each entry in the list of search results and deleted duplicate entries, entries with no corresponding keyphrases, and entries with clearly poor keyphrases. We were left with 90 distinct web pages with associated keyphrases. It is our impression that the web pages were typically submitted to Aliweb by their authors, so most of the keyphrases were probably supplied by the authors of the web pages.

Table 6 shows some statistics for the corpus. Note that the web page corpus does not have an internal structure, in the sense that the journal article corpus separates into five journals and the email corpus separates into six recipients. None of the keyphrases contain more than three words and 81% contain only one word. In this corpus, keyphrases tend to contain fewer words than in either the journal article corpus or the email corpus. About 70% of the keyphrases appear in the full text of the corresponding web pages.

Table 6: Some statistics for the Aliweb web page corpus.

| Description of Statistic | Value of Statistic |
|---|---|
| Average Number of Keyphrases per Document ± Standard Deviation | 6.0 ± 3.0 |
| Average Number of Words per Keyphrase ± Standard Deviation | 1.2 ± 0.5 |
| Average Number of Words per Document ± Standard Deviation | 949 ± 2603 |
| Percentage of the Keyphrases that Appear in the Full Text | 69.0% |

In the following experiments, this corpus is used for testing only. Thus there is no division of the data into testing and training sets. The keyphrases in this corpus seem subjectively to be of lower quality than the keyphrases in the other corpora. Although they are useful for testing purposes, they do not seem suitable for training.

## 5.4 The NASA Web Page Corpus

We collected 141 web pages from NASA’s Langley Research Center. Their Technology Applications Group (TAG) has 141 web pages that describe technology they have developed, available at http://tag-www.larc.nasa.gov/tops/tops_text.html. The web pages are intended to attract the interest of potential industrial partners and customers. Each page includes a list of keyphrases.

Table 7 shows that the documents are relatively short and relatively fewer keyphrases can be found in the bodies of the corresponding documents. This corpus has relatively more two- and three-word keyphrases than the other corpora. We used this corpus for testing only.

Table 7: Some statistics for the NASA web page corpus.

| Description of Statistic | Value of Statistic |
|---|---|
| Average Number of Keyphrases per Document ± Standard Deviation | 4.7 ± 2.0 |
| Average Number of Words per Keyphrase ± Standard Deviation | 1.9 ± 0.9 |
| Average Number of Words per Document ± Standard Deviation | 466 ± 102 |
| Percentage of the Keyphrases that Appear in the Full Text | 65.3% |

## 5.5 The FIPS Web Page Corpus

We gathered 35 web pages from the US government's Federal Information Processing Standards (FIPS), available at http://www.itl.nist.gov/div897/pubs/. These documents define the standards to which US government departments must conform when purchasing computer hardware and software. Each document includes a list of keyphrases.

From Table 8, we can see that the documents are relatively long and that many of the keyphrases appear in the body of the corresponding document. There is an unusual number of four-word keyphrases, because almost every document includes the keyphrase "Federal Information Processing Standard". If we ignore this phrase, the distribution is similar to the distribution in the Email Message corpus. This corpus was used for testing only.

Table 8: Some statistics for the FIPS web page corpus.

| Description of Statistic | Value of Statistic |
|---|---|
| Average Number of Keyphrases per Document ± Standard Deviation | 9.0 ± 3.5 |
| Average Number of Words per Keyphrase ± Standard Deviation | 2.0 ± 1.1 |
| Average Number of Words per Document ± Standard Deviation | 7025 ± 6385 |
| Percentage of the Keyphrases that Appear in the Full Text | 78.2% |

## 5.6 Testing and Training

We would like our learning algorithms to be able to perform well even when the testing data are significantly different from the training data. In a real-world application, it would be inconvenient if the learning algorithm required re-training for each new type of document. Therefore, our experiments do not use a random split of the documents into training and testing sets. Instead, we designed the experiments to test the ability of the learning algorithms to generalize to new data.

In our preliminary experiments, we found that the learning algorithms did generalize relatively well to new testing data. The main factor influencing the quality of the generalization appeared to be the average length of the documents in the training set, compared to the testing set. In a real-world application, it would be reasonable to have two different learned models, one for short documents and one for long documents. As Table 9 shows, we selected part of the journal article corpus to train the learning algorithms to handle long documents and part of the email message corpus to train the learning algorithms to handle short documents. During testing, we used the training corpus that was most similar to the given testing corpus, with respect to document lengths.

Table 9: The correspondence between testing and training data.

| Testing Corpus | | Corresponding Training Corpus | |
|---|---|---|---|
| Name | Number of Documents | Name | Number of Documents |
| Journal Articles — Testing Subset | 20 | Journal Article — Training Subset | 55 |
| Email Messages — Testing Subset | 76 | Email Messages — Training Subset | 235 |
| Aliweb Web Pages | 90 | Email Messages — Training Subset | 235 |
| NASA Web Pages | 141 | Email Messages — Training Subset | 235 |
| FIPS Web Pages | 35 | Journal Article — Training Subset | 55 |

# 6. *Applying C4.5 to Keyphrase Extraction*

In the first set of experiments, we used the C4.5 decision tree induction algorithm (Quinlan, 1993) to classify phrases as positive or negative examples of keyphrases. In this section, we describe the feature vectors, the settings we used for C4.5's parameters, the bagging procedure, and the method for sampling the training data.

## 6.1 Feature Vectors

We converted a document into a set of feature vectors by first making a list of all phrases of one, two, or three consecutive non-stop words that appear in the given document. (Stop words are words such as "the", "of", "and".) We used the Iterated Lovins stemmer (see Section 3.2) to find the stemmed form of each of these phrases. For each unique stemmed phrase, we generated a feature vector, as described in Table 10.

C4.5 has access to nine features (features 3 to 11) when building a decision tree. The leaves of the tree attempt to predict `class` (feature 12). When a decision tree predicts that the `class` of a vector is 1, then the phrase `whole_phrase` is a keyphrase, according to the tree. This phrase is suitable for output for a human reader. We used the stemmed form of the phrase, `stemmed_phrase`, for evaluating the performance of the tree. In our preliminary experiments, we evaluated 110 different features, before we settled on the features in Table 10.

Table 11 shows the number of feature vectors that were generated for each corpus. In total, we had more than 192,000 vectors for training and more than 168,000 vectors for testing. The large majority of these vectors were negative examples of keyphrases (class 0).

## 6.2 C4.5 Parameter Settings

In a real-world application, we assume that the user specifies the desired number of output keyphrases for a given document. However, a standard decision tree would not let the user control the number of feature vectors that are classified as belonging in class 1. Therefore we ran C4.5 with the `-p` option, which generates *soft-threshold* decision trees (Quinlan, 1993). Soft-threshold decision trees can generate a probability estimate for the class of each vector. For a given document, if the user specifies that *K* keyphrases are desired, then we select the *K* vectors that have the highest estimated probability of being in class 1.

Table 10: A description of the feature vectors used by C4.5.

| | Name of Feature | Description of Feature | C4.5 Type |
|---|---|---|---|
| 1 | `stemmed_phrase` | the stemmed form of a phrase<br>— for matching with human-generated phrases | ignore |
| 2 | `whole_phrase` | the most frequent whole (unstemmed) phrase corresponding to the given stemmed phrase<br>— for output and for calculating features 8 to 11 | ignore |
| 3 | `num_words_phrase` | the number of words in the phrase<br>— range: 1, 2, 3 | continuous |
| 4 | `first_occur_phrase` | the first occurrence of the stemmed phrase<br>— normalized by dividing by the number of words in the document (including stop words) | continuous |
| 5 | `first_occur_word` | the first occurrence of the earliest occurring single stemmed word in the stemmed phrase<br>— normalized by dividing by the number of words in the document (including stop words) | continuous |
| 6 | `freq_phrase` | the frequency of the stemmed phrase<br>— normalized by dividing by the number of words in the document (including stop words) | continuous |
| 7 | `freq_word` | the frequency of the most frequent single stemmed word in the stemmed phrase<br>— normalized by dividing by the number of words in the document (including stop words) | continuous |
| 8 | `relative_length` | the relative length of the most frequent whole phrase<br>— the number of characters in the whole phrase, normalized by dividing by the average number of characters in all candidate phrases | continuous |
| 9 | `proper_noun` | is the whole phrase a proper noun?<br>— based on the most frequent whole phrase | 0, 1 |
| 10 | `final_adjective` | does the whole phrase end in a final adjective?<br>— based on the most frequent whole phrase | 0, 1 |
| 11 | `common_verb` | does the whole phrase contain a common verb?<br>— based on the most frequent whole phrase | 0, 1 |
| 12 | `class` | is the stemmed phrase a keyphrase?<br>— based on match with stemmed form of human-generated keyphrases | 0, 1 |

In addition to the `-p` option, we also used `-c100` and `-m1`. The `-c` option sets the level of confidence for pruning and `-c100` results in minimal (but non-zero) pruning. The `-m` option sets the minimum number of examples that can be used to form a branch in the tree during training and `-m1` sets the minimum to one example. Compared to the default settings, these parameter settings result in relatively bushy trees, which tend to overfit the data. However, in our preliminary experiments, we found that bagging compensates for overfitting, so these parameter settings appear to work well when used in conjunction with bagging.

Table 11: The number of feature vectors for each corpus.

| Train/Test | Corpus Name | Number of Documents | Total Number of Vectors | Average Vectors Per Document | Percent Class 1 |
|---|---|---|---|---|---|
| Training | Journal | 55 | 158,240 | 2,877 | 0.20% |
| | Email | 235 | 34,378 | 146 | 2.44% |
| | All | 290 | 192,618 | 664 | 0.60% |
| Testing | Journal | 20 | 23,751 | 1,188 | 0.53% |
| | Email | 76 | 11,065 | 146 | 2.40% |
| | Aliweb | 90 | 26,752 | 297 | 1.08% |
| | NASA | 141 | 38,920 | 276 | 1.15% |
| | FIPS | 35 | 67,777 | 1,936 | 0.33% |
| | All | 362 | 168,265 | 465 | 0.80% |

### 6.3 Bagging Trees

Bagging involves generating many different decision trees and allowing them to vote on the classification of each example (Breiman, 1996a, 1996b; Quinlan, 1996). In general, decision tree induction algorithms have low bias but high variance. Bagging multiple trees tends to improve performance by reducing variance. Bagging appears to have relatively little impact on bias.

Because we used soft-threshold decision trees, we combined their probability estimates by averaging them, instead of voting. In preliminary experiments with the training documents, we obtained good results by bagging 50 decision trees. Adding more trees had no significant effect.

### 6.4 Sampling the Training Data

The standard approach to bagging is to randomly sample the training data, using sampling with replacement (Breiman, 1996a, 1996b; Quinlan, 1996). In preliminary experiments with the training data, we achieved good performance by training each of the 50 decision trees with a random sample of 1% of the training data.

The standard approach to bagging is to ignore the class when sampling, so the distribution of classes in the sample tends to correspond to the distribution in the training data as a whole. In Table 11, we see that the positive examples constitute only 0.2% to 2.4% of the total number of examples. To compensate for this, we modified the random sampling procedure so that 50% of the sampled examples were in class 0 and the other 50% were in class 1. This appeared to improve performance in preliminary experiments on the training data. This strategy is suggested in Kubat *et al.* (1998), where it is called *one-sided sampling*. Kubat *et al.* (1998) found that one-sided sampling significantly improved the performance of C4.5 on highly skewed data.

## 7. Experiment 1: Learning to Extract Keyphrases with C4.5

This section presents four experiments with C4.5. In Experiment 1A, we establish a baseline for the performance of C4.5, using the configuration described in Section 6. We bag 50 trees, generated by randomly sampling 1% of the training data, with equal numbers of samples from the two classes (keyphrase and non-keyphrase). In Experiment 1B, we vary the number of trees. The results support the hypothesis that 50 trees are better than one tree. In Experiment 1C, we vary the ratio of the classes. The results do not support the hypothesis that one-sided sampling improves performance. In Experiment 1D, we vary the size of the random samples. The results confirm the hypothesis that 1% sampling is better than 50% sampling.

## 7.1 Experiment 1A: The Baseline Algorithm

In the baseline configuration of C4.5, we bag 50 trees, where each tree is trained on a random sample of 1% of the training data, with equal samples from both classes. The performance is measured by the precision when the desired number of phrases is set to 5, 7, 9, 11, 13, and 15. The precision is measured separately for each document in the given corpus, and then the average precision is calculated for the corpus.

Figure 6 shows the baseline performance of C4.5. The plots show the precision for the testing data only (see Table 9). It appears that the e-mail model generalizes relatively well to the Aliweb and NASA corpora, but the journal model does not generalize well to the FIPS corpus.

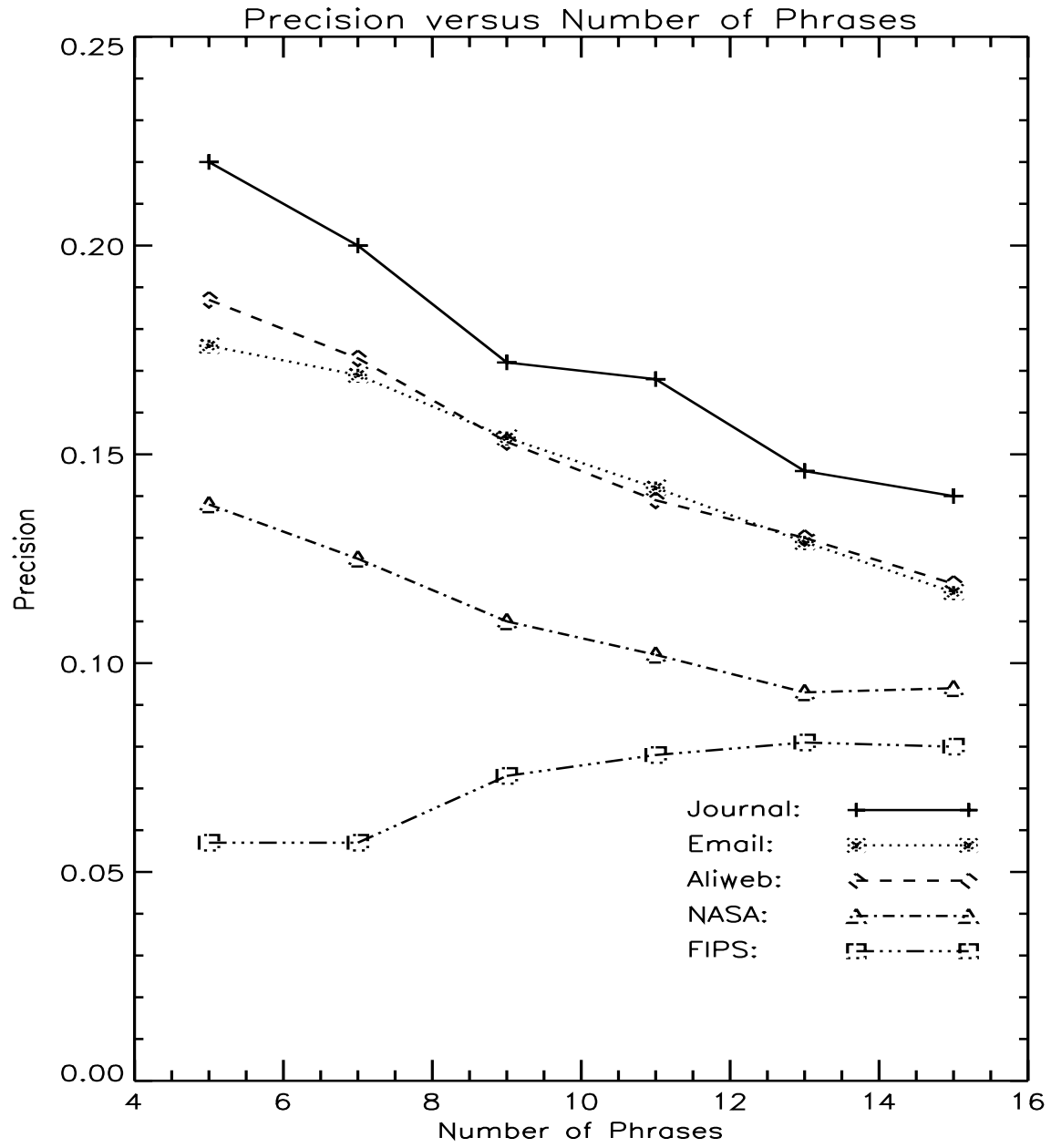

Figure 6: Experiment 1A: The baseline precision of C4.5 at various cut-offs for the desired number of extracted keyphrases.

Table 12 shows the time required to train and Table 13 shows the time required to test, measured in seconds. The time in parentheses is the average time per document. Training involves (1) generating feature vectors from the documents, (2) randomly sampling the training data, and (3) building the decision trees. Table 12 shows the time for each of these three steps. Testing involves (1) generating feature vectors from the documents and (2) using the decision trees. Table 13 shows the time for each of these two steps. All of the code was written in C and executed on a Pentium II 233 running Windows NT 4.0.[7]

Table 14 shows the phrases selected by the baseline configuration of C4.5 for three articles from *Psycoloquy* (i.e., the journal article testing documents). In these three examples, the desired number of phrases is set to nine. The phrases in bold match the author's phrases, according to the Iterated Lovins stemming algorithm (see Section 3).

Table 12: Experiment 1A: Training time for the baseline configuration of C4.5.

| Corpus Name | Number of Documents | Total Time (Average Time) in Seconds | | | |
|---|---|---|---|---|---|
| | | Make Vectors | Random Selection | Make 50 Trees | Total |
| Journal | 55 | 56 (1.0) | 147 (2.7) | 47 (0.9) | 250 (4.5) |
| Email | 235 | 32 (0.1) | 34 (0.1) | 23 (0.1) | 89 (0.4) |

Table 13: Experiment 1A: Testing time for the baseline configuration of C4.5.

| Corpus Name | Number of Documents | Total Time (Average Time) in Seconds | | |
|---|---|---|---|---|
| | | Make Vectors | Use 50 Trees | Total |
| Journal | 20 | 9 (0.5) | 28 (1.4) | 37 (1.9) |
| Email | 76 | 11 (0.1) | 95 (1.3) | 106 (1.4) |
| Aliweb | 90 | 17 (0.2) | 113 (1.3) | 130 (1.4) |
| NASA | 141 | 24 (0.2) | 176 (1.2) | 200 (1.4) |
| FIPS | 35 | 28 (0.8) | 58 (1.7) | 86 (2.5) |

Table 14: Experiment 1A: Examples of the selected phrases for three articles from *Psycoloquy*.

| | |
|---|---|
| Title: | "The Base Rate Fallacy Myth" |
| Author's Keyphrases: | base rate fallacy, Bayes' theorem, decision making, ecological validity, ethics, fallacy, judgment, probability. |
| C4.5's Top Nine Keyphrases: | **judgments**, base rates, **base rate fallacy**, **decision making**, posteriors, **fallacy**, **probability**, rate fallacy, probabilities. |
| Precision: | 0.556 |
| Title: | "Brain Rhythms, Cell Assemblies and Cognition: Evidence from the Processing of Words and Pseudowords" |
| Author's Keyphrases: | brain theory, cell assembly, cognition, event related potentials, ERP, electroencephalograph, EEG, gamma band, Hebb, language, lexical processing, magnetoencephalography, MEG, psychophysiology, periodicity, power spectral analysis, synchrony. |
| C4.5's Top Nine Keyphrases: | cell assemblies, **cognitive**, responses, assemblies, cognitive processing, brain functions, word processing, oscillations, cell. |
| Precision: | 0.111 |
| Title: | "On the Evolution of Consciousness and Language" |
| Author's Keyphrases: | consciousness, language, plans, motivation, evolution, motor system. |
| C4.5's Top Nine Keyphrases: | psychology, **language**, **consciousness**, behavior, **evolution**, cognitive psychology, Bridgeman, organization, modern cognitive psychology. |
| Precision: | 0.333 |

## 7.2 Experiment 1B: Varying the Number of Trees

This experiment tests the hypothesis that bagging improves the performance of C4.5 on the task of automatic keyphrase extraction. Figure 7 shows the precision when the desired number of phrases is set to 5, 7, 9, 11, 13, and 15. The number of trees is set to 1, 25, and 50. For four of the corpora, the precision tends to rise as the number of trees increases. The exception is the FIPS corpus. As we noted in the previous section, C4.5 has difficulty in generalizing from the journal article training data to the FIPS testing data.

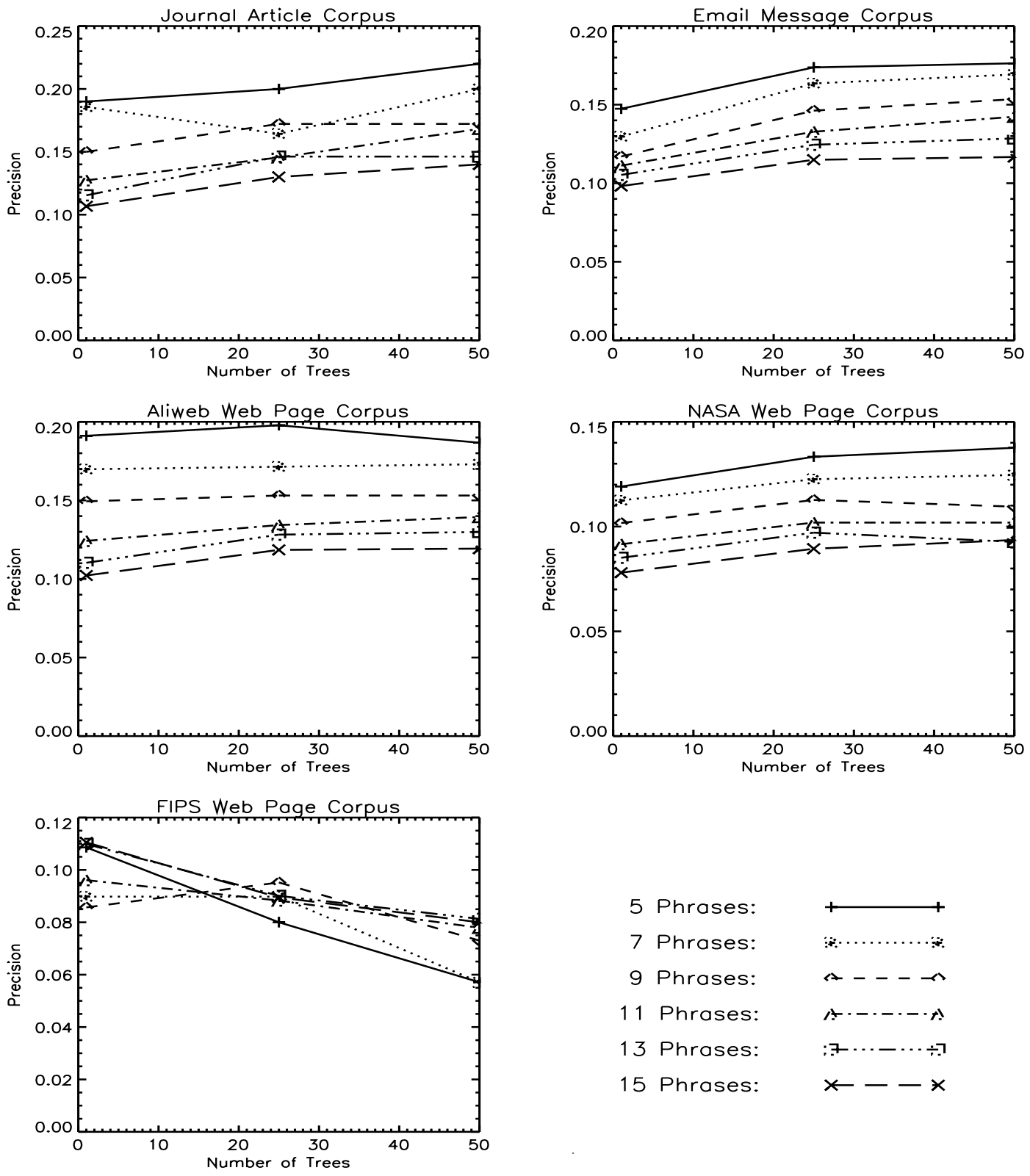

Figure 7: Experiment 1B: The effect of varying the number of trees on precision.

In Table 15, we test the significance of this rising trend, using a paired t-test. The table shows that, when we look at the five testing collections together, 50 trees are significantly more precise than 1 tree, when the desired number of phrases is set to 15. The only case in which 50 trees are significantly worse than 1 tree is with the FIPS collection.

## 7.3 Experiment 1C: Varying the Ratios of the Classes

This experiment tests the hypothesis that one-sided sampling (Kubat *et al.*, 1998) can help C4.5 handle the skewed class distribution. Figure 8 shows the precision when the percentage of examples in class 1 (positive examples of keyphrases) is set to 1%, 25%, and 50%. For at least three of the corpora, precision tends to fall as the percentage increases.

Table 16 shows that, when we look at the five testing collections together, there is a significant drop in precision when 50% of the samples are positive examples, compared to 1%. Only the email collection appears to benefit from balanced sampling of the classes. We must reject the hypothesis that one-sided sampling (Kubat *et al.*, 1998) is useful for our data. Although our preliminary experiments with the training data suggested that one-sided sampling would be benefi-cial, the hypothesis is not supported by the testing data.

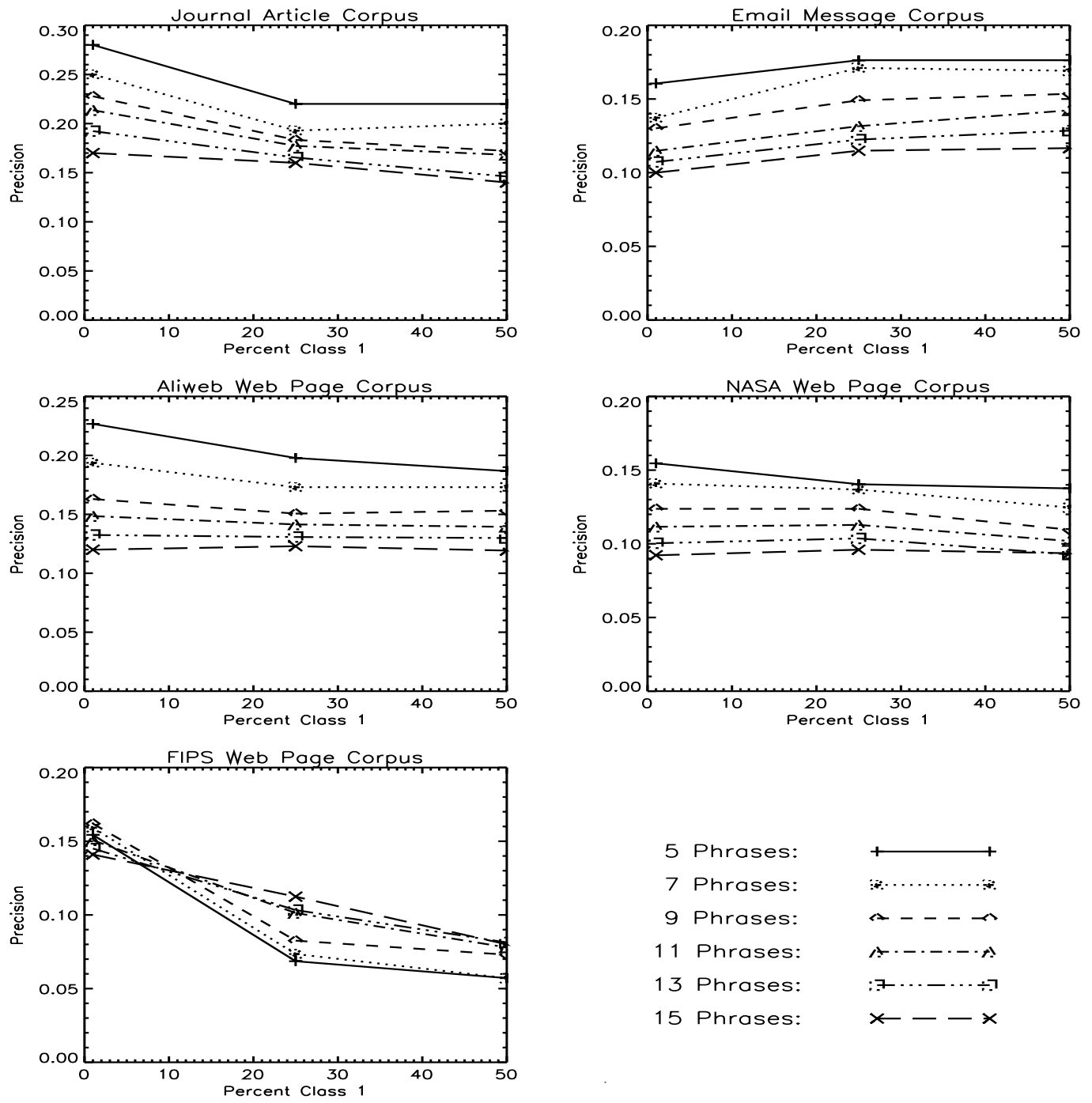

Figure 8: Experiment 1C: The effect of varying the percentage of class 1 on precision.

Table 15: Experiment 1B: A comparison of 50 trees with 1 tree.

| Corpus Name | Number of Documents | Number of Phrases | Average Precision ± Standard Deviation | | | Significant with 95% Confidence |
|---|---|---|---|---|---|---|
| | | | 1 Tree | 50 Trees | 50 - 1 | |
| Journal | 20 | 5 | 0.190 ± 0.229 | 0.220 ± 0.182 | 0.030 ± 0.218 | NO |
| | | 15 | 0.107 ± 0.098 | 0.140 ± 0.078 | 0.033 ± 0.085 | NO |
| Email | 76 | 5 | 0.147 ± 0.151 | 0.176 ± 0.160 | 0.029 ± 0.141 | NO |
| | | 15 | 0.098 ± 0.079 | 0.117 ± 0.099 | 0.018 ± 0.060 | YES |
| Aliweb | 90 | 5 | 0.191 ± 0.182 | 0.187 ± 0.166 | -0.004 ± 0.164 | NO |
| | | 15 | 0.102 ± 0.076 | 0.119 ± 0.082 | 0.017 ± 0.057 | YES |
| NASA | 141 | 5 | 0.119 ± 0.137 | 0.138 ± 0.129 | 0.018 ± 0.133 | NO |
| | | 15 | 0.078 ± 0.066 | 0.094 ± 0.069 | 0.016 ± 0.048 | YES |
| FIPS | 35 | 5 | 0.109 ± 0.101 | 0.057 ± 0.092 | -0.051 ± 0.148 | YES |
| | | 15 | 0.111 ± 0.067 | 0.080 ± 0.062 | -0.031 ± 0.078 | YES |
| All | 362 | 5 | 0.146 ± 0.158 | 0.155 ± 0.151 | 0.009 ± 0.151 | NO |
| | | 15 | 0.093 ± 0.074 | 0.106 ± 0.080 | 0.013 ± 0.060 | YES |

Table 16: Experiment 1C: A comparison of 1% positive examples with 50% positive examples.

| Corpus Name | Number of Documents | Number of Phrases | Average Precision ± Standard Deviation | | | Significant with 95% Confidence |
|---|---|---|---|---|---|---|
| | | | 1% Class 1 | 50% Class 1 | 50 - 1 | |
| Journal | 20 | 5 | 0.280 ± 0.255 | 0.220 ± 0.182 | -0.060 ± 0.216 | NO |
| | | 15 | 0.170 ± 0.113 | 0.140 ± 0.078 | -0.030 ± 0.103 | NO |
| Email | 76 | 5 | 0.161 ± 0.160 | 0.176 ± 0.160 | 0.016 ± 0.145 | NO |
| | | 15 | 0.100 ± 0.081 | 0.117 ± 0.099 | 0.017 ± 0.055 | YES |
| Aliweb | 90 | 5 | 0.227 ± 0.190 | 0.187 ± 0.166 | -0.040 ± 0.135 | YES |
| | | 15 | 0.120 ± 0.074 | 0.119 ± 0.082 | -0.001 ± 0.048 | NO |
| NASA | 141 | 5 | 0.155 ± 0.159 | 0.138 ± 0.129 | -0.017 ± 0.138 | NO |
| | | 15 | 0.092 ± 0.068 | 0.094 ± 0.069 | 0.001 ± 0.045 | NO |
| FIPS | 35 | 5 | 0.154 ± 0.162 | 0.057 ± 0.092 | -0.097 ± 0.184 | YES |
| | | 15 | 0.141 ± 0.066 | 0.080 ± 0.062 | -0.061 ± 0.063 | YES |
| All | 362 | 5 | 0.181 ± 0.177 | 0.155 ± 0.151 | -0.026 ± 0.151 | YES |
| | | 15 | 0.110 ± 0.078 | 0.106 ± 0.080 | -0.004 ± 0.058 | NO |

### 7.4 Experiment 1D: Varying the Sizes of the Samples

This experiment tests the hypothesis that sampling 1% of the training data results in better precision than larger samples. Figure 9 shows the precision when the sample size is 1%, 25%, and 50%. For three of the copora, increasing the sample size tends to decrease the precision. The exceptions are the email message corpus and the FIPS web page corpus.

In Table 17, we test the significance of this trend, using a paired t-test. The table shows that, when we look at the five testing collections together, a 1% sample rate yields better precision than a 50% sample rate, when the desired number of phrases is set to 15. This supports the hypothesis that a relatively small sample size is better for bagging than a large sample. This is expected, since bagging works best when the combined models are heterogeneous (Breiman, 1996a, 1996b; Quinlan, 1996). Increasing the sample size tends to make the models more homogenous.

## 8. *GenEx: A Hybrid Genetic Algorithm for Keyphrase Extraction*

We have experimented with many ways of applying C4.5 to automatic keyphrase extraction. The preceding section presented a few of these experiments. During the course of our experimentation, we came to believe that a tailor-made algorithm for learning to extract keyphrases might be able to achieve better precision than a general-purpose learning algorithm such as C4.5. This motivated us to develop the *GenEx* algorithm.

GenEx has two components, the *Genitor* genetic algorithm (Whitley, 1989) and the *Extractor* keyphrase extraction algorithm (NRC, patent pending). Extractor takes a document as input and produces a list of keyphrases as output. Extractor has twelve parameters that determine how it processes the input text. In GenEx, the parameters of Extractor are tuned by the Genitor genetic algorithm (Whitley, 1989), to maximize performance (fitness) on training data. Genitor is used to tune Extractor, but Genitor is no longer needed once the training process is complete. When we

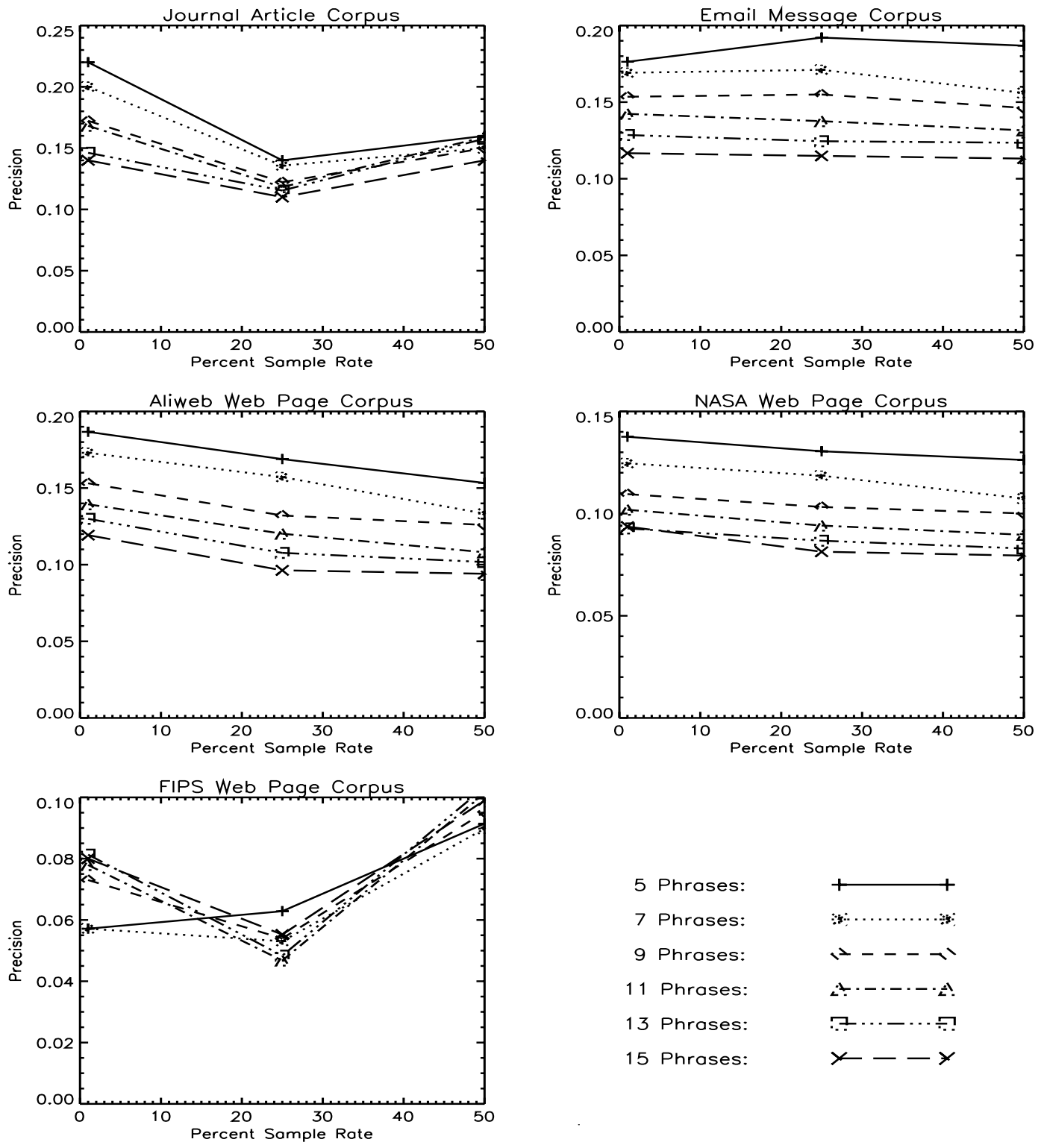

Figure 9: Experiment 1D: The effect of varying the sample rate on precision.

Table 17: Experiment 1D: A comparison of 1% sample rate with 50% sample rate.

| Corpus Name | Number of Documents | Number of Phrases | Average Precision ± Standard Deviation | | | Significant with 95% Confidence |
|---|---|---|---|---|---|---|
| | | | 1% Sample Rate | 50% Sample Rate | 50 - 1 | |
| Journal | 20 | 5 | 0.220 ± 0.182 | 0.160 ± 0.139 | -0.060 ± 0.131 | NO |
| | | 15 | 0.140 ± 0.078 | 0.140 ± 0.094 | 0.000 ± 0.061 | NO |
| Email | 76 | 5 | 0.176 ± 0.160 | 0.187 ± 0.161 | 0.011 ± 0.146 | NO |
| | | 15 | 0.117 ± 0.099 | 0.113 ± 0.102 | -0.004 ± 0.062 | NO |
| Aliweb | 90 | 5 | 0.187 ± 0.166 | 0.153 ± 0.144 | -0.033 ± 0.168 | NO |
| | | 15 | 0.119 ± 0.082 | 0.094 ± 0.071 | -0.025 ± 0.064 | YES |
| NASA | 141 | 5 | 0.138 ± 0.129 | 0.126 ± 0.136 | -0.011 ± 0.143 | NO |
| | | 15 | 0.094 ± 0.069 | 0.079 ± 0.057 | -0.014 ± 0.055 | YES |
| FIPS | 35 | 5 | 0.057 ± 0.092 | 0.091 ± 0.112 | 0.034 ± 0.133 | NO |
| | | 15 | 0.080 ± 0.062 | 0.099 ± 0.059 | 0.019 ± 0.050 | YES |
| All | 362 | 5 | 0.155 ± 0.151 | 0.144 ± 0.144 | -0.010 ± 0.150 | NO |
| | | 15 | 0.106 ± 0.080 | 0.095 ± 0.076 | -0.011 ± 0.060 | YES |

know the best parameter values, we can discard Genitor. Thus the learning system is called

GenEx (Genitor plus Extractor) and the trained system is called Extractor (GenEx minus Genitor).

## 8.1 Extractor

What follows is a conceptual description of the Extractor algorithm. For clarity, we describe Extractor at an abstract level that ignores efficiency considerations. That is, the actual Extractor software is essentially an efficient implementation of the following algorithm.[8]

There are ten steps to the Extractor algorithm. Figure 10 summarizes the ten steps. Steps 4 and 5 are conceptually independent of steps 1, 2, and 3, so they are represented as a separate sequence. (For efficiency reasons, in the actual implementation of the algorithm, several steps are interleaved.) Table 18 is a list of the 12 parameters of Extractor, with a brief description of each of them. The meaning of the parameters should become clear as the algorithm is described.

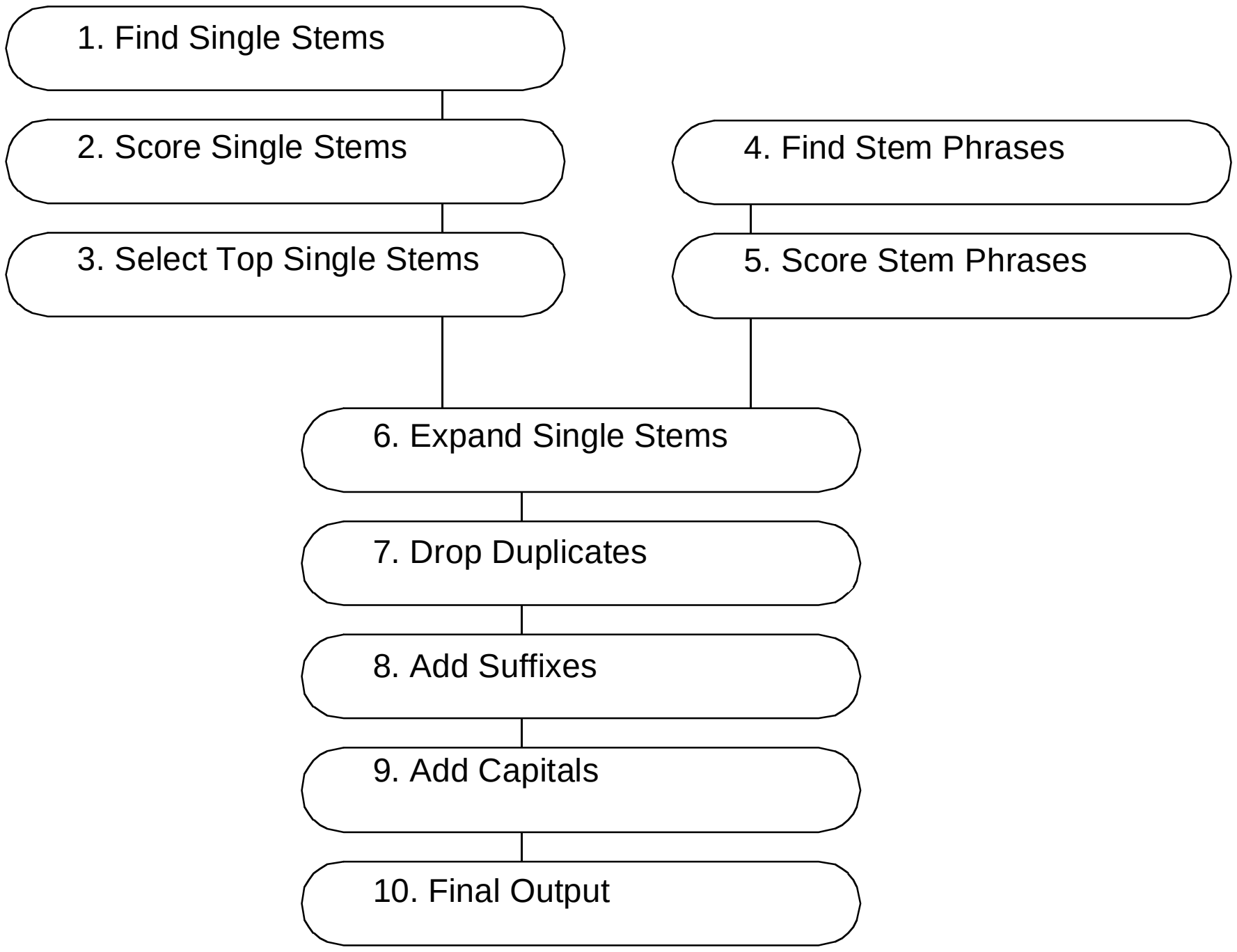

Figure 10: An overview of the Extractor algorithm.

The ten steps of the algorithm are:

1. **Find Single Stems:** Make a list of all of the words in the input text. Drop words with less than three characters. Drop stop words (words like “and”, “or”, “if”, “he”, “she”), using a given stop word list. Convert all remaining words to lower case. Stem the words by truncating them at STEM_LENGTH characters.

The advantages of this simple form of stemming (stemming by truncation) are speed and flexibility. Stemming by truncation is much faster than either the Lovins (1968) or Porter (1980) stemming algorithms. The aggressiveness of the stemming can be adjusted by changing STEM_LENGTH. When STEM_LENGTH is low (e.g., five characters), stemming by truncation is more aggressive than Lovins or Porter stemming. When STEM_LENGTH is high (e.g., ten charac-ters), truncation is less aggressive. This gives Genitor control over the level of aggressiveness.

Table 18: The twelve parameters of Extractor, with some sample values.

| Parameter Number | Parameter Name | Sample Value | Description |
|---|---|---|---|
| 1 | NUM_PHRASES | 10 | length of final phrase list |
| 2 | NUM_WORKING | 50 | length of working list |
| 3 | FACTOR_TWO_ONE | 2.33 | factor for expanding to two words |
| 4 | FACTOR_THREE_ONE | 5.00 | factor for expanding to three words |
| 5 | MIN_LENGTH_LOW_RANK | 0.9 | low rank words must be longer than this |
| 6 | MIN_RANK_LOW_LENGTH | 5 | short words must rank higher than this |
| 7 | FIRST_LOW_THRESH | 40 | definition of “early” occurrence |
| 8 | FIRST_HIGH_THRESH | 400 | definition of “late” occurrence |
| 9 | FIRST_LOW_FACTOR | 2.0 | reward for “early” occurrence |
| 10 | FIRST_HIGH_FACTOR | 0.65 | penalty for “late” occurrence |
| 11 | STEM_LENGTH | 5 | maximum characters for fixed length stem |
| 12 | SUPPRESS_PROPER | 0 | flag for suppressing proper nouns |

2. **Score Single Stems:** For each unique stem, count how often the stem appears in the text and note when it first appears. If the stem “evolut” first appears in the word “Evolution”, and “Evolution” first appears as the tenth word in the text, then the first appearance of “evolut” is said to be in position 10. Assign a score to each stem. The score is the number of times the stem appears in the text, multiplied by a factor. If the stem first appears before FIRST_LOW_THRESH, then multiply the frequency by FIRST_LOW_FACTOR. If the stem first appears after FIRST_HIGH_THRESH, then multiply the frequency by FIRST_HIGH_FACTOR.

Typically FIRST_LOW_FACTOR is greater than one and FIRST_HIGH_FACTOR is less than one. Thus, early, frequent stems receive a high score and late, rare stems receive a low score. This gives Genitor control over the weight of early occurrence versus the weight of frequency. Early occurrence and frequency are both good clues that a stem is important, but it is not obvious how the two clues should be combined.

3. **Select Top Single Stems:** Rank the stems in order of decreasing score and make a list of the top NUM_WORKING single stems.

Cutting the list at NUM_WORKING, as opposed to allowing the list to have an arbitrary length, improves the efficiency of Extractor. It also acts as a filter for eliminating lower quality stems.

4. **Find Stem Phrases:** Make a list of all phrases in the input text. A phrase is defined as a sequence of one, two, or three words that appear consecutively in the text, with no intervening stop words or phrase boundaries (punctuation characters). Stem each phrase by truncating each word in the phrase at STEM_LENGTH characters.

In our corpora, phrases of four or more words are relatively rare. Therefore Extractor only considers phrases of one, two, or three words. Extractor does not consider phrases with embedded stop words, because authors tend to avoid embedded stop words in their keyphrases. For example, instead of “shift of bias”, which contains the stop word “of”, an author would prefer “bias shift” as a keyphrase.

5. **Score Stem Phrases:** For each stem phrase, count how often the stem phrase appears in the text and note when it first appears. Assign a score to each phrase, exactly as in step 2, using the parameters FIRST_LOW_FACTOR, FIRST_LOW_THRESH, FIRST_HIGH_FACTOR, and FIRST_HIGH_THRESH. Then make an adjustment to each score, based on the number of stems in the phrase. If there is only one stem in the phrase, do nothing. If there are two stems in the phrase, multiply the score by FACTOR_TWO_ONE. If there are three stems in the phrase, multiply the score by FACTOR_THREE_ONE.

Typically FACTOR_TWO_ONE and FACTOR_THREE_ONE are greater than one, so this adjustment will increase the score of longer phrases. A phrase of two or three stems is necessarily never more frequent than the most frequent single stem contained in the phrase. The factors FACTOR_TWO_ONE and FACTOR_THREE_ONE are designed to boost the score of longer phrases, to compensate for the fact that longer phrases are expected to otherwise have lower scores than shorter phrases.

6. **Expand Single Stems:** For each stem in the list of the top NUM_WORKING single stems, find the highest scoring stem phrase of one, two, or three stems that contains the given single stem. The result is a list of NUM_WORKING stem phrases. Keep this list ordered by the scores calculated in step 2.

Now that the single stems have been expanded to stem phrases, we no longer need the scores that were calculated in step 5. That is, the score for a stem phrase (step 5) is now replaced by the score for its corresponding single stem (step 2). The reason is that the adjustments to the score that were introduced in step 5 are useful for expanding the single stems to stem phrases, but they are not useful for comparing or ranking stem phrases.

7. **Drop Duplicates:** The list of the top NUM_WORKING stem phrases may contain duplicates. For example, two single stems may expand to the same two-word stem phrase. Delete duplicates from the ranked list of NUM_WORKING stem phrases, preserving the highest ranked phrase.

For example, suppose that the stem “evolu” (e.g., “evolution” truncated at five characters) appears in the fifth position in the list of the top NUM_WORKING single stems and “psych” (e.g., “psychology” truncated at five characters) appears in the tenth position. When the single stems are expanded to stem phrases, we might find that “evolu psych” (e.g., “evolutionary psychology” truncated at five characters) appears in the fifth and tenth positions in the list of stem phrases. In this case, we delete the phrase in the tenth position. If there are duplicates, then the list now has fewer than NUM_WORKING stem phrases.

8. **Add Suffixes:** For each of the remaining stem phrases, find the most frequent corresponding whole phrase in the input text. For example, if “evolutionary psychology” appears ten times in the text and “evolutionary psychologist” appears three times, then “evolutionary psychology” is the more frequent corresponding whole phrase for the stem phrase “evolu psych”. When counting the frequency of whole phrases, if a phrase has an ending that indicates a possible adjective, then the frequency for that whole phrase is set to zero. An ending such as “al”, “ic”, “ible”, etc., indicates a possible adjective. Adjectives in the middle of a phrase (for example, the second word in a three-word phrase) are acceptable; only phrases that end in adjectives are penalized. Also, if a phrase contains a verb, the frequency for that phrase is set to zero. To check for verbs, we use a list of common verbs. A word that might be either a noun or a verb is included in this list only when it is much more common for the word to appear as a verb than as a noun.

For example, suppose the input text contains “manage”, “managerial”, and “management”. If STEM_LENGTH is, say, five, the stem "manag" will be expanded to "management" (a noun), because the frequency of "managerial" will be set to zero (because it is an adjective, ending in "al") and the frequency of "manage" will be set to zero (because it is a verb, appearing in the list of common verbs). Although "manage" and "managerial" would not be output, their presence in the input text helps to boost the score of the stem "manag" (as measured in step 2), and thereby increase the likelihood that "management" will be output.

9. **Add Capitals:** For each of the whole phrases (phrases with suffixes added), find the best capitalization, where *best* is defined as follows. For each word in a phrase, find the capitalization with the least number of capitals. For a one-word phrase, this is the best capitalization. For a two-word or three-word phrase, this is the best capitalization, unless the capitalization is inconsistent. The capitalization is said to be inconsistent when one of the words has the capitalization pattern of a proper noun but another of the words does not appear to be a proper noun (e.g., "Turing test"). When the capitalization is inconsistent, see whether it can be made consistent by using the capitalization with the second lowest number of capitals (e.g., "Turing Test"). If it cannot be made consistent, use the inconsistent capitalization. If it can be made consistent, use the consistent capitalization.

For example, given the phrase "psychological association", the word "association" might appear in the text only as "Association", whereas the word "psychological" might appear in the text as "PSYCHOLOGICAL", "Psychological", and "psychological". Using the least number of capitals, we get "psychological Association", which is inconsistent. However, it can be made consistent, as "Psychological Association".

10. **Final Output:** We now have an ordered list of mixed-case (upper and lower case, if appropriate) phrases with suffixes added. The list is ordered by the scores calculated in step 2. That is, the score of each whole phrase is based on the score of the highest scoring single stem that appears in the phrase. The length of the list is at most NUM_WORKING, and is likely less, due to step 7. We now form the final output list, which will have at most NUM_PHRASES phrases. We go through the list of phrases in order, starting with the top-ranked phrase, and output each phrase that passes the following tests, until either NUM_PHRASES phrases have been output or we reach the end of the list. The tests are (1) the phrase should not have the capitalization of a proper noun, unless the flag SUPPRESS_PROPER is set to 0 (if 0 then allow proper nouns; if 1 then suppress proper nouns); (2) the phrase should not have an ending that indicates a possible adjective; (3) the phrase should be longer than MIN_LENGTH_LOW_RANK, where the length is measured by the ratio of the number of characters in the candidate phrase to the number of characters in the average phrase, where the average is calculated for all phrases in the input text that consist of one to three consecutive non-stop words; (4) if the phrase is shorter than MIN_LENGTH_LOW_RANK, it may still be acceptable, if its rank in the list of candidate phrases is better than (closer to the top of the list than) MIN_RANK_LOW_LENGTH; (5) if the phrase fails both tests (3) and (4), it may still be acceptable, if its capitalization pattern indicates that it is probably an abbreviation; (6) the phrase should not contain any words that are most commonly used as verbs; (7) the phrase should not match any phrases in a given list of stop phrases (where "match" means equal strings, ignoring case, but including suffixes).

That is, a phrase must pass tests (1), (2), (6), (7), and at least one of tests (3), (4), and (5).

Although our experimental procedure does not consider capitalization or suffixes when comparing machine-generated keyphrases to human-generated keyphrases, steps 8 and 9 are still use-ful, because some of the screening tests in step 10 are based on capitalization and suffixes. Of course, steps 8 and 9 are essential when the output is intended for human readers.

## 8.2 Genitor

A genetic algorithm may be viewed as a method for optimizing a string of bits, using techniques that are inspired by biological evolution. A genetic algorithm works with a set of bit strings, called a *population* of *individuals*. The initial population is usually randomly generated. New individuals (new bit strings) are created by randomly changing existing individuals (this operation is called *mutation*) and by combining substrings from *parents* to make new *children* (this operation is called *crossover*). Each individual is assigned a score (called its *fitness*) based on some measure of the quality of the bit string, with respect to a given task. Fitter individuals get to have more children than less fit individuals. As the genetic algorithm runs, new individuals tend to be increasingly fit, up to some asymptote.

Genitor is a *steady-state* genetic algorithm (Whitley, 1989), in contrast to many other genetic algorithms, such as Genesis (Grefenstette 1983, 1986), which are *generational*.[9] A generational genetic algorithm updates its entire population in one batch, resulting in a sequence of distinct generations. A steady-state genetic algorithm updates its population one individual at a time, resulting in a continuously changing population, with no distinct generations. Typically a new individual replaces the least fit individual in the current population. Whitley (1989) suggests that steady-state genetic algorithms tend to be more aggressive (they have greater selective pressure) than generational genetic algorithms.

In some preliminary tests, we compared Genitor (Whitley, 1989) and Genesis (Grefenstette 1983, 1986) for tuning Extractor. They gave similar average performance, but Genitor appeared to have lower variance in its performance across repeated runs, so we chose Genitor for the experiments reported here. However, we did not make a thorough comparison of the two genetic algorithms, using different mutation rates and population sizes, since this is not the focus of our work. Thus we do not claim that we have demonstrated that Genitor is superior to Genesis for tuning Extractor.

## 8.3 GenEx: Genitor Plus Extractor

The parameters in Extractor are set using the standard machine learning paradigm of supervised learning. The algorithm is tuned with a dataset, consisting of documents paired with target lists of keyphrases. The dataset is divided into training and testing subsets. The learning process involves adjusting the parameters to maximize the match between the output of Extractor and the target keyphrase lists, using the training data. The success of the learning process is measured by examining the match using the testing data.

We assume that the user sets the value of NUM_PHRASES, the desired number of phrases, to a value between five and fifteen. We then set NUM_WORKING to $5 \cdot$ NUM_PHRASES. The remaining ten parameters are set by Genitor. Genitor uses a binary string of 72 bits to represent the ten parameters, as shown in Table 19. We run Genitor with a population size of 50 for 1050 trials (these are default settings). Each trial consists of running Extractor with the parameter settings specified in the given binary string, processing the entire training set. The fitness measure for the binary string is based on the average precision for the whole training set. The final output of Genitor is the highest scoring binary string. Ties are broken by choosing the earlier string.

We first tried to use the average precision on the training set as the fitness measure, but GenEx discovered that it could achieve high average precision by adjusting the parameters so that less than NUM_PHRASES phrases were output. This is clearly not desirable, so we modified the fitness measure to penalize GenEx when less than NUM_PHRASES phrases were output:

$$\text{total_matches} = \text{total number of matches between GenEx and human} \tag{5}$$

$$\text{total_machine_phrases} = \text{total number of phrases output by GenEx} \tag{6}$$

$$\text{precision} = \text{total_matches} / \text{total_machine_phrases} \tag{7}$$

$$\text{num_docs} = \text{number of documents in training set} \tag{8}$$

$$\text{total_desired} = \text{num_docs} \cdot \text{NUM_PHRASES} \tag{9}$$

$$\text{penalty} = (\text{total_machine_phrases} / \text{total_desired})^2 \tag{10}$$

$$\text{fitness} = \text{precision} \cdot \text{penalty} \tag{11}$$

Table 19: The ten parameters of Extractor that are tuned by Genitor, with types and ranges.

| Parameter Number | Parameter Name | Parameter Type | Parameter Range | Number of Bits |
|---|---|---|---|---|
| 3 | FACTOR_TWO_ONE | real | [1, 3] | 8 |
| 4 | FACTOR_THREE_ONE | real | [1, 5] | 8 |
| 5 | MIN_LENGTH_LOW_RANK | real | [0.3, 3.0] | 8 |
| 6 | MIN_RANK_LOW_LENGTH | integer | [1, 20] | 5 |
| 7 | FIRST_LOW_THRESH | integer | [1, 1000] | 10 |
| 8 | FIRST_HIGH_THRESH | integer | [1, 4000] | 12 |
| 9 | FIRST_LOW_FACTOR | real | [1, 15] | 8 |
| 10 | FIRST_HIGH_FACTOR | real | [0.01, 1.0] | 8 |
| 11 | STEM_LENGTH | integer | [1, 10] | 4 |
| 12 | SUPPRESS_PROPER | boolean | [0, 1] | 1 |
| Total Number of Bits in Binary String: | | | | 72 |

The penalty factor varies between 0 and 1. It has no effect (i.e., it is 1) when the number of phrases output by GenEx equals the desired number of phrases. The penalty grows (i.e., it approaches 0) with the square of the gap between the desired number of phrases and the actual number of phrases. Preliminary experiments on the training data confirmed that this fitness measure led GenEx to find parameter values with high average precision while ensuring that NUM_PHRASES phrases were output.

Since STEM_LENGTH is modified by Genitor during the GenEx learning process, the fitness measure used by Genitor is not based on stemming by truncation. If the fitness measure were based on stemming by truncation, a change in STEM_LENGTH would change the apparent fitness, even if the actual output keyphrase list remained constant. Therefore fitness is measured with the Iterated Lovins stemmer.

We ran Genitor with a Selection Bias of 2.0 and a Mutation Rate of 0.2. These are the default settings for Genitor. We used the Adaptive Mutation operator and the Reduced Surrogate Crossover operator (Whitley, 1989). Adaptive Mutation determines the appropriate level of mutation for a child according to the hamming distance between its two parents; the less the difference, the higher the mutation rate. Reduced Surrogate Crossover first identifies all positions in which the parent strings differ. Crossover points are only allowed to occur in these positions.

## 8.4 Comparison of C4.5 with GenEx

A comparison of Extractor (Section 8.1) with the feature vectors we used with C4.5 (Section 6.1) shows that GenEx and C4.5 are learning with essentially the same feature sets. The two algorithms have access to the same information, but they learn different kinds of models of keyphrases. This section lists some of the more significant differences between GenEx and C4.5 (as we have used it here).

1. Given a set of phrases with a shared single-word stem (for example, the set of phrases {"learning", "machine learning", "learnability"} shares the single-word stem "learn"), GenEx tends to choose the best member of the set, rather than choosing the whole set. GenEx first identifies the shared single-word stem (steps 1 to 3 in Section 8.1) and then looks for the best representative phrase in the set (steps 4 to 6). C4.5 tends to choose several members from the set, if it chooses any of them.[10]
2. GenEx can adjust the aggressiveness of the stemming, by adjusting STEM_LENGTH. C4.5 must take the stems that are given in the training data.[11]
3. C4.5 is designed to yield high accuracy. GenEx is designed to yield high precision for a given NUM_PHRASES. High precision does not necessarily correspond to high accuracy.
4. C4.5 uses the same model (the same set of decision trees) for all values of NUM_PHRASES. With C4.5, we select the top NUM_PHRASES most probable feature vectors, but our estimate of probability is not sensitive to the value of NUM_PHRASES. On the other hand, Genitor tunes Extractor differently for each desired value of NUM_PHRASES.
5. GenEx might output less than the desired number of phrases, NUM_PHRASES, but C4.5 (as we use it here) always generates exactly NUM_PHRASES phrases. Therefore, in the following experiments, performance is measured by the average precision, where precision is defined by (12), not by (13). Equation (12) ensures that GenEx cannot spuriously boost its score by generating fewer phrases than the user requests.[12]

$$\text{precision} = \text{number of matches} \,/\, \text{desired number of machine-generated phrases} \tag{12}$$

$$\text{precision} = \text{number of matches} \,/\, \text{actual number of machine-generated phrases} \tag{13}$$

# 9. *Experiment 2: Learning to Extract Keyphrases with GenEx*

This set of experiments compares GenEx to C4.5. In Figure 11, we compare GenEx to both the baseline configuration of C4.5 (Experiment 1A) and the best configuration of C4.5 (Experiment 1C). It is not quite fair to use the best configuration, because we only know that it is the best by looking at the testing data, but GenEx does not have the advantage of any information from the testing data. However, the performance of GenEx is good enough that we can afford to be generous to C4.5.

Figure 11 shows the average precision on each testing corpus, with the desired number of phrases set to 5, 7, 9, 11, 13, and 15. In Table 20, we test the significance of the difference between GenEx and the best configuration of C4.5. The table shows that, when we look at the five testing collections together, GenEx is significantly more precise. There is no case in which the performance of GenEx is below the performance of C4.5.

Table 21 shows the training time for GenEx and Table 22 shows the testing time. GenEx was trained separately for each value of NUM_PHRASES, 5, 7, 9, 11, 13, and 15. The fourth column in Table 21 shows the average training time for a single value of NUM_PHRASES. In comparison with C4.5, GenEx is much slower to train (Table 12), but also much faster after it has been trained (Table 13). (The same computer was used for timing C4.5 and GenEx.)

# 9. *Experiment 2: Learning to Extract Keyphrases with GenEx*

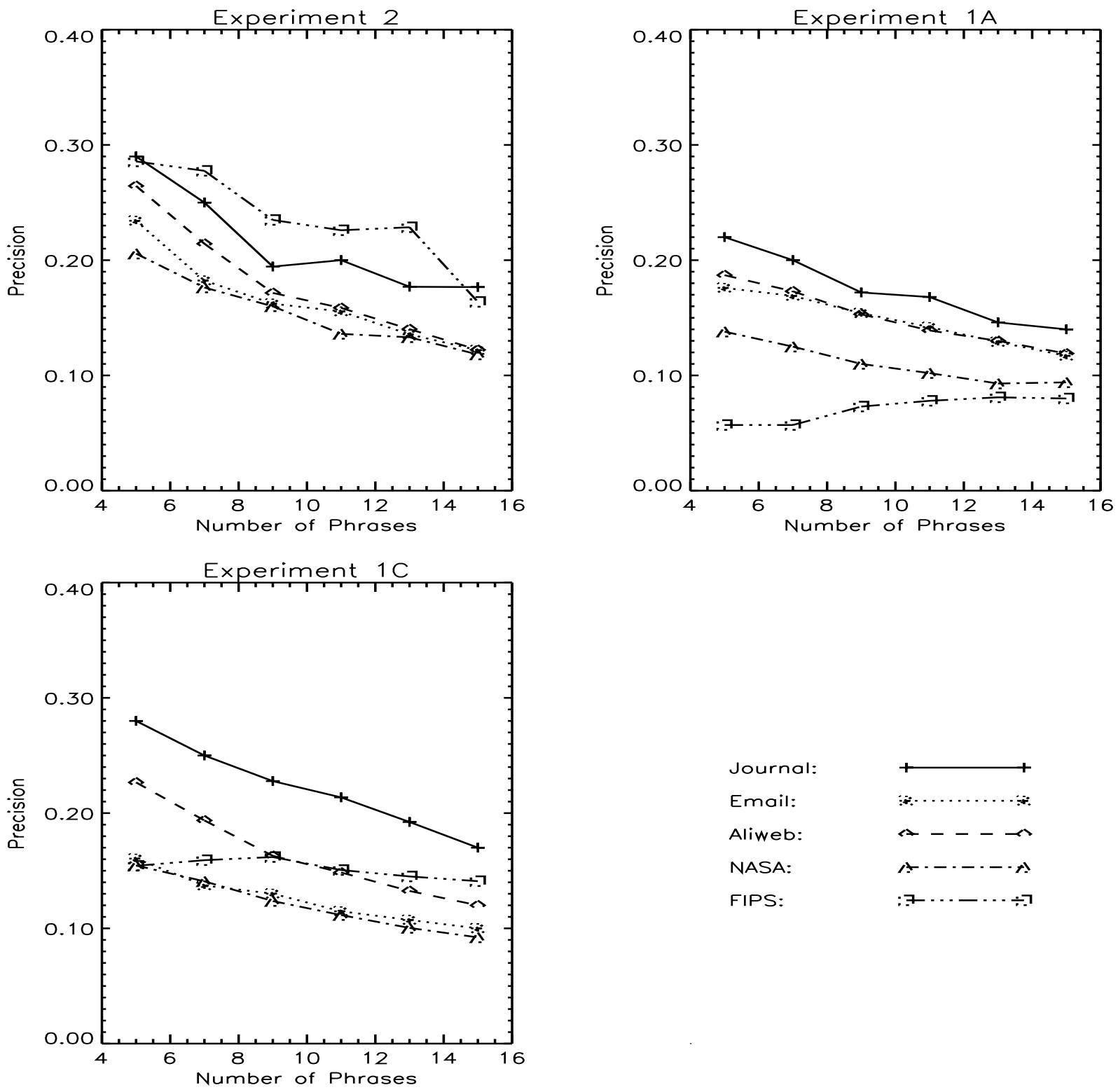

Figure 11: A comparison of GenEx (Experiment 2) with C4.5 (Experiments 1A and 1C).

Table 20: A comparison of GenEx (Experiment 2) with C4.5 (Experiment 1C).

| Corpus Name | Number of Documents | Number of Phrases | Average Precision ± Standard Deviation | | | Significant with 95% Confidence |
|---|---|---|---|---|---|---|
| | | | GenEx | C4.5 (1C) | GenEx - C4.5 | |
| Journal | 20 | 5 | 0.290 ± 0.247 | 0.280 ± 0.255 | 0.010 ± 0.137 | NO |
| | | 15 | 0.177 ± 0.130 | 0.170 ± 0.113 | 0.007 ± 0.061 | NO |
| Email | 76 | 5 | 0.234 ± 0.205 | 0.161 ± 0.160 | 0.074 ± 0.166 | YES |
| | | 15 | 0.122 ± 0.105 | 0.100 ± 0.081 | 0.022 ± 0.073 | YES |
| Aliweb | 90 | 5 | 0.264 ± 0.177 | 0.227 ± 0.190 | 0.038 ± 0.185 | NO |
| | | 15 | 0.122 ± 0.077 | 0.120 ± 0.074 | 0.002 ± 0.077 | NO |
| NASA | 141 | 5 | 0.206 ± 0.172 | 0.155 ± 0.159 | 0.051 ± 0.136 | YES |
| | | 15 | 0.118 ± 0.080 | 0.092 ± 0.068 | 0.026 ± 0.068 | YES |
| FIPS | 35 | 5 | 0.286 ± 0.170 | 0.154 ± 0.162 | 0.131 ± 0.222 | YES |
| | | 15 | 0.164 ± 0.078 | 0.141 ± 0.066 | 0.023 ± 0.081 | NO |
| All | 362 | 5 | 0.239 ± 0.187 | 0.181 ± 0.177 | 0.058 ± 0.167 | YES |
| | | 15 | 0.128 ± 0.089 | 0.110 ± 0.078 | 0.018 ± 0.073 | YES |

Table 23 presents some examples of the phrases selected by GenEx, when NUM_PHRASES is set to nine. Matches with the authors (according to the Iterated Lovins stemming algorithm) are in bold.

Table 21: Training time for GenEx.

| Corpus Name | Number of Documents | Time in Hours:Minutes:Seconds | |
|---|---|---|---|
| | | Total | Average Per Given Number of Phrases |
| Journal | 55 | 48:28:03 | 08:04:40 |
| Email | 235 | 14:54:15 | 02:29:02 |

Table 22: Testing time for GenEx.

| Corpus Name | Number of Documents | Time in Seconds | |
|---|---|---|---|
| | | Total | Average Per Document |
| Journal | 20 | 5 | 0.25 |
| Email | 76 | 4 | 0.05 |
| Aliweb | 90 | 6 | 0.07 |
| NASA | 141 | 8 | 0.06 |
| FIPS | 35 | 12 | 0.34 |

Table 23: Experiment 2: Examples of the selected phrases for three articles from *Psycoloquy*.

| | |
|---|---|
| Title: | "The Base Rate Fallacy Myth" |
| Author's Keyphrases: | base rate fallacy, Bayes' theorem, decision making, ecological validity, ethics, fallacy, judgment, probability. |
| GenEx's Top Nine Keyphrases: | base rates, **judgments**, **probability**, decision, **base rate fallacy**, prior, experiments, **decision making**, probabilistic information. |
| Precision: | 0.444 |
| Title: | "Brain Rhythms, Cell Assemblies and Cognition: Evidence from the Processing of Words and Pseudowords" |
| Author's Keyphrases: | brain theory, cell assembly, cognition, event related potentials, ERP, electroencephalograph, EEG, gamma band, Hebb, language, lexical processing, magnetoencephalography, MEG, psychophysiology, periodicity, power spectral analysis, synchrony. |
| GenEx's Top Nine Keyphrases: | neurons, pseudowords, responses, cell assemblies, ignition, activation, brain, cognitive processing, gamma-band responses. |
| Precision: | 0.000 |
| Title: | "On the Evolution of Consciousness and Language" |
| Author's Keyphrases: | consciousness, language, plans, motivation, evolution, motor system. |
| GenEx's Top Nine Keyphrases: | **plans**, **consciousness**, **language**, planning, psychology, behavior, memory, cognitive psychology, plan-executing. |
| Precision: | 0.333 |

# 10. *Experiment 3: Keyphrases for Metadata: Comparison of Word 97 Metadata with GenEx Metadata*

It is not clear what it means to say, for example, that GenEx has a precision between 20% and 30% when the desired number of phrases is five. What does this mean in practical terms? Our answer to this question is to compare GenEx with commercial software. In this section, we compare GenEx and Microsoft's Word 97, when applied to metadata generation (see Section 2.1).

## 10.1 Microsoft's Word 97: The AutoSummarize Feature

Microsoft's Word 97 is a complete word processing software package. In this paper, we are only concerned with the AutoSummarize feature in Word 97. The AutoSummarize feature is available from the *Tools* menu. The main function of this feature is to identify important sentences in the document that is being edited. The identified sentences can be highlighted or separated from the remainder of the text. The user can specify a target percentage of the text for AutoSummarize to mark as important.

As a side-effect, when AutoSummarize is used, it also fills in the *Keywords* field of the document's *Properties*. The *Properties* form is available from the *File* menu. AutoSummarize always generates exactly five keyphrases (if the document contains at least five distinct words). The keyphrases are always single words, never phrases of two or more words. They are always in lower case, even when they are abbreviations or proper nouns. There is no way for the user of Word 97 to adjust the AutoSummarize feature. For example, it is not possible to ask it for six keyphrases instead of five.

## 10.2 Experiment

The histogram in Figure 12 compares the precision of Word 97 and GenEx, when the desired number of phrases is set to five. Table 24 shows that GenEx is significantly more precise, when we look at the five testing collections together. GenEx has higher precision on all five corpora and significantly higher precision on three of the corpora. Table 25 presents the Word 97 metadata for three sample documents, with matches in bold.

Table 24: Experiment 3: A comparison of Word 97 and GenEx on the metadata task.

| Corpus Name | Number of Documents | Number of Phrases | Average Precision ± Standard Deviation | | | Significant with 95% Confidence |
|---|---|---|---|---|---|---|
| | | | Word 97 | GenEx | Word 97 - GenEx | |
| Journal | 20 | 5 | 0.170 ± 0.187 | 0.290 ± 0.247 | -0.120 ± 0.151 | YES |
| Email | 76 | 5 | 0.145 ± 0.137 | 0.234 ± 0.205 | -0.089 ± 0.192 | YES |
| Aliweb | 90 | 5 | 0.236 ± 0.198 | 0.264 ± 0.177 | -0.029 ± 0.196 | NO |
| NASA | 141 | 5 | 0.084 ± 0.138 | 0.206 ± 0.172 | -0.122 ± 0.174 | YES |
| FIPS | 35 | 5 | 0.246 ± 0.195 | 0.286 ± 0.170 | -0.040 ± 0.180 | NO |
| All | 362 | 5 | 0.155 ± 0.175 | 0.239 ± 0.187 | -0.084 ± 0.186 | YES |

# 11. *Experiment 4: Keyphrases for Highlighting: Comparison of Search 97 Highlighting with GenEx Highlighting*

This section continues the theme of the previous section, the practical implications of GenEx's

# 11. *Experiment 4: Keyphrases for Highlighting: Comparison of*

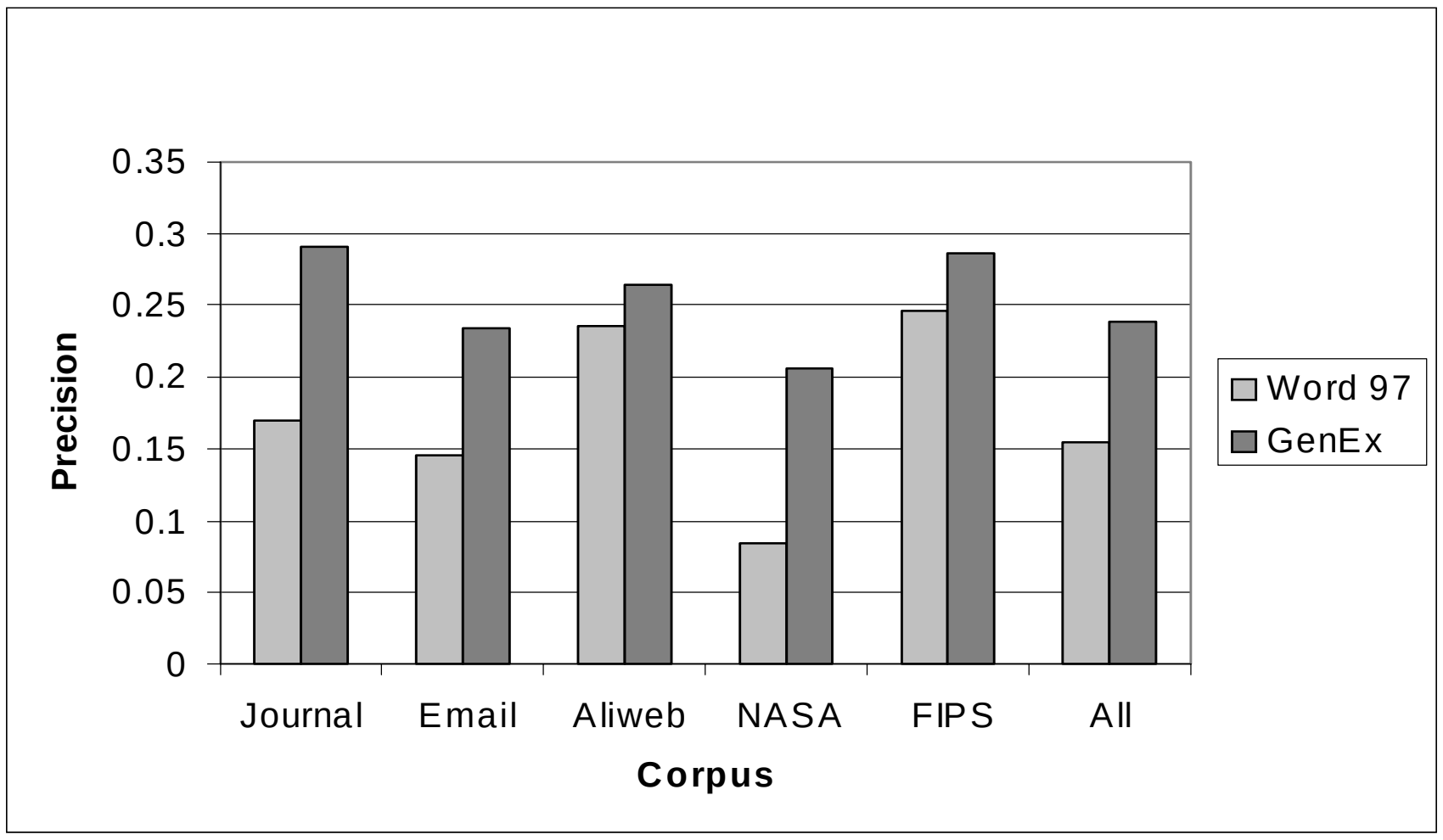

Figure 12: Experiment 3: A comparison of Word 97 and GenEx on the metadata task.

Table 25: Experiment 3: Examples of the selected phrases for three articles from *Psycoloquy*.

| | |
|---|---|
| Title: | "The Base Rate Fallacy Myth" |
| Author's Keyphrases: | base rate fallacy, Bayes' theorem, decision making, ecological validity, ethics, fallacy, judgment, probability. |
| Word 97's Top Five Keyphrases: | rate, base, information, **judgment**, psychology. |
| Precision: | 0.200 |
| Title: | "Brain Rhythms, Cell Assemblies and Cognition: Evidence from the Processing of Words and Pseudowords" |
| Author's Keyphrases: | brain theory, cell assembly, cognition, event related potentials, ERP, electroencephalograph, EEG, gamma band, Hebb, language, lexical processing, magnetoencephalography, MEG, psychophysiology, periodicity, power spectral analysis, synchrony. |
| Word 97's Top Five Keyphrases: | assembly, neuron, word, activity, process. |
| Precision: | 0.000 |
| Title: | "On the Evolution of Consciousness and Language" |
| Author's Keyphrases: | consciousness, language, plans, motivation, evolution, motor system. |
| Word 97's Top Five Keyphrases: | **plan**, **consciousness**, process, **language**, action. |
| Precision: | 0.600 |

performance. Here we compare GenEx and Verity's Search 97, when applied to highlighting (see Section 2.2).

## 11. *Experiment 4: Keyphrases for Highlighting: Comparison of*

### 11.1 Verity's Search 97: The Summarization Feature

Verity's Search 97 is a complete text retrieval system, including a search engine, an index builder, and a web crawler. In this paper, we are only concerned with the Summarization feature in Search 97. The Summarization feature enables the search engine to display summaries of each document in the hit list, along with the usual information, such as the document title. A summary in Search 97 consists of a list of sentences, with highlighted keyphrases embedded in the sentences. The user can control the number of sentences in a summary, by either specifying the number of sentences desired or the percentage of the source document desired.

### 11.2 Experiment

In these experiments, we use Search 97 to automatically generate summaries with 2, 4, and 6 sentences each. For each summary, we list the phrases that are automatically highlighted by Search 97. When two or more phrases in the list match (according to the Iterated Lovins stemmer), we keep only one of the phrases.[13] These are the *highlight lists* for Search 97.

To make the highlight lists for GenEx, we intersect the summaries of Search 97 with the keyphrases extracted by GenEx, when the desired number of phrases is set to 15. That is, for each of the fifteen phrases extracted by GenEx, we see whether the phrase appears in the corresponding summary generated by Search 97. We say that a phrase appears in a summary if the stems of the words in the phrase match the stems of a consecutive sequence of words in the summary.

The numbers of phrases in the highlight lists are not under our direct control, although we can partially control them indirectly by varying the number of sentences in the summaries. Therefore, in these experiments, we use the F-measure instead of precision. Figure 13 plots the F-measures of the Search 97 and GenEx highlight lists, as a function of the number of sentences in the summaries. Table 26 shows that the F-measure for GenEx is significantly higher than the F-measure for Search 97, when we look at the five testing collections together. Table 27 shows the highlight lists for a sample document.

Table 26: A comparison of Search 97 and GenEx on the highlighting task.

| Corpus Name | Number of Documents | Number of Sentences | Average F-measure ± Standard Deviation | | | Significant with 95% Confidence |
|---|---|---|---|---|---|---|
| | | | Search 97 | GenEx | Search 97 - GenEx | |
| Journal | 20 | 2 | 0.133 ± 0.141 | 0.205 ± 0.199 | -0.073 ± 0.169 | NO |
| | | 6 | 0.164 ± 0.123 | 0.244 ± 0.182 | -0.080 ± 0.133 | YES |
| Email | 76 | 2 | 0.197 ± 0.242 | 0.232 ± 0.228 | -0.035 ± 0.219 | NO |
| | | 6 | 0.169 ± 0.132 | 0.219 ± 0.175 | -0.050 ± 0.128 | YES |
| Aliweb | 90 | 2 | 0.187 ± 0.193 | 0.186 ± 0.163 | 0.001 ± 0.158 | NO |
| | | 6 | 0.154 ± 0.122 | 0.180 ± 0.139 | -0.027 ± 0.098 | YES |
| NASA | 141 | 2 | 0.124 ± 0.147 | 0.180 ± 0.149 | -0.056 ± 0.156 | YES |
| | | 6 | 0.109 ± 0.091 | 0.179 ± 0.129 | -0.070 ± 0.107 | YES |
| FIPS | 35 | 2 | 0.158 ± 0.107 | 0.191 ± 0.110 | -0.033 ± 0.106 | NO |
| | | 6 | 0.189 ± 0.111 | 0.207 ± 0.113 | -0.018 ± 0.105 | NO |
| All | 362 | 2 | 0.157 ± 0.181 | 0.193 ± 0.172 | -0.036 ± 0.168 | YES |
| | | 6 | 0.142 ± 0.116 | 0.192 ± 0.145 | -0.050 ± 0.112 | YES |

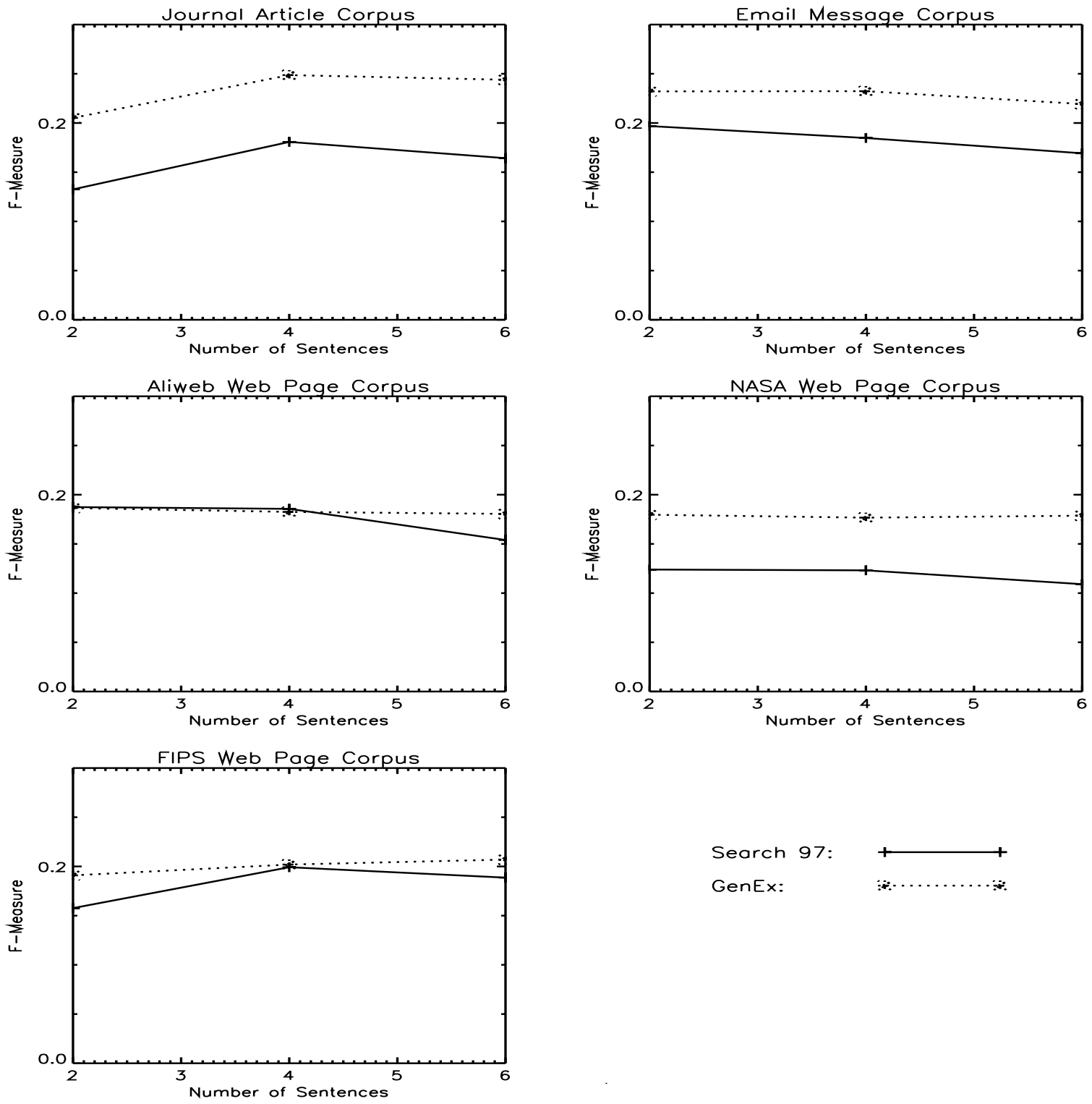

Figure 13: Experiment 4: A comparison of Search 97 and GenEx on the highlighting task, with varying numbers of sentences.

# 12. *Discussion*

This section discusses the current status of Extractor and our plans for future work.

## 12.1 Current Status of Extractor

We have licensed the Extractor component of GenEx to Tetranet Software Inc. and to ThinkTank Technologies Corp.[14] Tetranet Software is using Extractor in their Wisebot product and ThinkTank Technologies is using Extractor in their Virtual Research Assistant product.[15] To facilitate embedding Extractor within other software systems, Extractor is written as a DLL (Dynamically Linked Library) with an API (Application Program Interface). The API is fully reentrant, to allow several documents to be processed simultaneously, using separate threads for each document. Multithreading is useful, for example, when processing web pages.

Extractor 3.1 (the current version, at the time of writing) works with English and French documents. It has, built into it, four sets of parameter values, for long and short documents in English and French. The algorithm for French documents is almost the same as the algorithm for English documents (as described in Section 8.1). Extractor automatically detects the language of the document and applies the appropriate algorithm and parameter values.

## 12.2 Future Work

We are currently adding Japanese and German capability to Extractor. We have just begun Ger-man, but we have a working prototype of Japanese. The algorithm for Japanese is significantly different from the algorithm for English and French, although there is much in common at an abstract level.

Table 27: An example of the performance of Search 97 and GenEx on the highlighting task.

| | |
|---|---|
| Title: | "On the Evolution of Consciousness and Language" |
| Author's Keyphrases: | consciousness, language, plans, motivation, evolution, motor system. |
| Number of Sentences: | 6 |
| Search 97 Summary: | Psychology can be based on plans, internally held images of achievement that organize the stimulus-response links of traditional psychology. Consciousness is the operation of the plan-executing mechanism, enabling behavior to be driven by plans rather than immediate environmental contingencies.The mechanism unpacks a single internally held idea into a series of actions. Language comprehension uses the plan-monitoring mechanism to pack a series of linguistic events into an idea. Recursive processing results from monitoring one's own speech. A new psychology of plans promises to include consciousness by organizing and synthesizing the many subdisciplines that have grown within psychology and the cognitive neurosciences. |
| Search 97 Highlights: | psychology, **plans**, organize, **consciousness**, plan-executing mechanism, behavior, mechanism, idea, actions, **language**, processing. |
| Search 97 Performance: | Precision: 0.273, Recall: 0.500, F-measure: 0.353. |
| GenEx Highlights: | **plans**, **consciousness**, **language**, behavior, psychology, organization, processes, environment. |
| GenEx Performance: | Precision: 0.375, Recall: 0.500, F-measure: 0.429. |

In the future, we plan to add the capability to select important sentences, to automatically generate a summary of a document. We also plan to add a mechanism for recognizing synonyms, to improve the performance of the keyphrase extraction.

## 13. *Conclusion*

In this paper, we have presented two approaches to the task of learning to extract keyphrases from text. The first approach was to apply the C4.5 decision tree induction algorithm (Quinlan, 1993), using bagging (Breiman, 1996a, 1996b; Quinlan, 1996) and one-sided sampling (Kubat *et al.*, 1998). The experiments support the claim that bagging is helpful for this task, but one-sided sampling is not. Our experience with C4.5 led us to suspect that a custom-designed learning algorithm might perform better than the general-purpose C4.5 algorithm. We presented the GenEx algorithm and experiments that support the claim that GenEx performs better than C4.5.

The practical implications of the first two sets of experiments are not obvious. The practical question is whether the machine-generated keyphrases are good enough to be useful. One way to answer this question might be to ask people for their subjective opinions about the quality of the keyphrases. For example, we could mix machine-generated phrases with human-generated phrases and ask people to select their favourites. However, such a survey would not necessarily answer the question. The machine-generated phrases might be very useful, even if they are not as good as human-generated keyphrases. Extractor can make a keyphrase list for an average journal article in one quarter of a second (Table 22). The speed of automatic keyphrase extraction makes it possible to use keyphrases in applications where it would not be economically feasible to use human-generated keyphrases, even if they were subjectively superior to machine-generated keyphrases.

Our test of the practical value of GenEx was to compare it to commercial tools that perform keyphrase extraction, for metadata generation and highlighting. The third set of experiments supports the claim that GenEx is superior to Word 97's algorithm for generating keyphrase metadata. The fourth set of experiments supports the claim that GenEx is superior to Search 97's algorithm for highlighting. Neither of these commercial algorithms use machine learning techniques, so the experiments lend some support to the claim that machine learning is a valuable approach to automatic keyphrase extraction.

## *Acknowledgments*

Thanks to Elaine Sin of the University of Calgary for creating the keyphrases for the email message corpus. Thanks to Joel Martin of the NRC, for writing the Java applet that is illustrated in Figure 3. Thanks to my colleagues at NRC and the University of Ottawa for their support, encouragement, and constructive criticism.

## *Notes*

1. Microsoft and Word 97 are trademarks or registered trademarks of Microsoft Corporation. Verity and Search 97 are trademarks or registered trademarks of Verity Inc.
2. We used an implementation of the Porter (1980) stemming algorithm written in Perl, by Jim Richardson, at the University of Sydney, Australia. This implementation includes some extensions to Porter's original algorithm, to handle British spelling. It is available at http://www.maths.usyd.edu.au:8000/jimr.html. For the Lovins (1968) stemming algorithm, we used an implementation written in C, by Linh Huynh. This implementation is part of the MG (Managing Gigabytes) search engine, which was developed by a group of people in Australia and New Zealand. The MG code is available at http://www.kbs.citri.edu.au/mg/.
3. There may be some exceptions, such as the use of the phrase "the Mob" to refer to an international crime organization.
4. Some journals ask their authors to order their keyphrases from most general to most specific. In this paper, we have ignored the order of the keyphrases. For most of the applications we have considered here, the order is not important. GenEx attempts to order the keyphrases it produces from most important to least important.
5. It is now available as a commercial product, called NetOwl Extractor, from IsoQuest. See http://www.isoquest.com/.
6. The INSPEC database is the leading English bibliographic database for scientific and technical literature in physics, electrical engineering, electronics, communications, control engineering, computers and computing, and information technology. It is produced by the Institution of Electrical Engineers. Records in the INSPEC database have fields for both controlled vocabulary index terms (called descriptors) and free index terms (called identifiers). More information is available at http://www.iee.org.uk/publish/inspec/inspec.html.
7. The decision tree routines were written by Quinlan (1993). We wrote the feature vector generation routines and the random sampling routines. The code was carefully written for speed.
8. Extractor is written in C. A demonstration version of Extractor is available at http://ai.iit.nrc.ca/II_public/extractor.html. The demonstration version has been trained already; it does not allow the user to make any adjustments.

# 13. *Conclusion*

9. The source code for Genitor is available at ftp://ftp.cs.colostate.edu/pub/GENITOR.tar. The source code for Genesis is available at ftp://www.aic.nrl.navy.mil/pub/galist/src/genesis.tar.Z. Both of these programs are written in C.
10. It would be possible to add a post-processing step that winnows the phrases selected by C4.5, but this would be a step down the path that leads from the general-purpose C4.5 algorithm to the custom-made GenEx algorithm. Our point is that a custom-made algorithm has advantages over a general-purpose algorithm.
11. Again, it would be possible to adjust the stemming procedure externally, by cross-validation, but this is another step down the path that leads from C4.5 to GenEx.
12. If GenEx does generate fewer than num_phrases phrases, then we could randomly select further phrases until we have num_phrases phrases. Thus (12) does not misrepresent the performance of GenEx. Note that we do not use the fitness measure (11) to evaluate the performance of GenEx.
13. Since the phrases match each other, they would match the same human-generated keyphrase, so there is no need to keep more than one in the list.
14. See http://www.tetranetsoftware.com/ and http://www.thinktanktech.com/.
15. Tetranet and Wisebot are trademarks or registered trademarks of Tetranet Software. ThinkTank and Virtual Research Assistant are trademarks or registered trademarks of ThinkTank Technologies. Extractor is an Official Mark of the National Research Council of Canada.